\documentclass[11pt]{article}

\usepackage[margin=1in]{geometry} \usepackage[T1]{fontenc} \usepackage{mathpazo} \usepackage{courier} \usepackage{microtype}  \usepackage[skip=0.55\baselineskip plus 2pt]{parskip} \usepackage{float}
\usepackage{authblk}
\usepackage{placeins}   

  \usepackage[sort]{natbib} \usepackage{hyperref} \usepackage{url} \usepackage{bm} \usepackage{multirow} \usepackage{booktabs} \usepackage{axlib}

\definecolor{utorange}{HTML}{BF5700} \definecolor{stdnavy}{HTML}{1A4F9C} \colorlet{axaccent}{black} \usepackage{titlesec} \titleformat{\section} {\Large\bfseries}{\textcolor{axaccent}{\thesection}}{0.9em}{} \titleformat{\subsection} {\large\bfseries}{\textcolor{axaccent}{\thesubsection}}{0.9em}{} \titleformat{\subsubsection} {\normalsize\bfseries}{\textcolor{axaccent}{\thesubsubsection}}{0.9em}{} \hypersetup{colorlinks=true, linkcolor=stdnavy, citecolor=stdnavy, urlcolor=stdnavy}

\title{\dqfam{}: Hybrid-Attention Diffusion Language Models}

\author{Anton Xue \qquad Litu Rout \qquad Aditya Akella \\[2pt]
Adam Klivans \qquad Sujay Sanghavi \qquad Sanjay Shakkottai}

\affil{University of Texas at Austin}
\affil{\small\texttt{\{anton.xue, litu.rout, sanjay.shakkottai\}@utexas.edu}, \\
  \texttt{\{akella, klivans\}@cs.utexas.edu},\; \texttt{sanghavi@mail.utexas.edu}}

\date{}

\begin{document}

\maketitle

\begin{abstract}
Adapting a pretrained autoregressive (AR) model is a cost-efficient route to a diffusion language model (DLM).
While nearly all such adaptations start from a full-attention transformer, AR modeling has shifted toward hybrid architectures that interleave attention and RNN layers.
This creates an obstacle for adaptation: unlike attention, RNNs are structurally causal and nontrivial to bidirectionalize.
Despite this mismatch, we investigate whether such backbones can become effective DLMs by adapting Qwen3.5 at 0.8B, 2B, 4B, and 9B scales, yielding the dQwen3.5 family.
We find that hybrid backbones can be efficient starting points for adaptation: against a full-attention control, the hybrid reaches a given training loss in about half the tokens.
Across scales, dQwen3.5 resembles full-attention DLMs in any-order decoding behavior and performs strongly under parallel decoding.
\end{abstract}

\newcommand{\artifactlogo}[3]{%
  \raisebox{-#1\height}{\includegraphics[height=#2]{#3}}%
}

\begin{center}
\small
\artifactlogo{0.35}{1.18em}{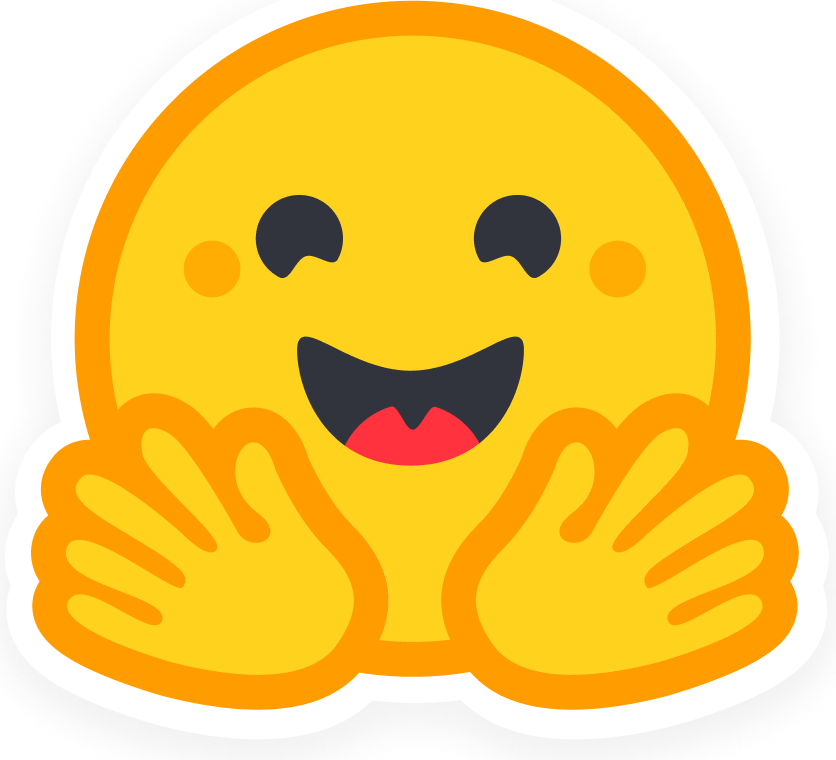}\hspace{0.42em}%
\href{https://huggingface.co/UT-IFML/dQwen3.5-9B-Base}{\texttt{huggingface.co/UT-IFML/dQwen3.5-9B-Base}}
\end{center}

\section{Introduction}
\label{sec:intro}

Diffusion language models (DLMs) enable parallel and any-order generation by iteratively unmasking positions rather than strictly decoding left to right~\citep{austin2021structured,shi2024simplified,sahoo2024simple}.
While recent releases are increasingly competitive with autoregressive (AR) models \citep{nie2025large,ye2025dream}, training a DLM from scratch remains expensive: reaching parity with a similarly sized AR model can require an order of magnitude more tokens~\citep{nie2024scaling}.
AR-to-DLM adaptation has thus emerged as a cheaper alternative for training a DLM: one initializes from a pretrained AR model, makes its causal attention layers bidirectional, and continues training with a diffusion objective~\citep{cheng2025sdar,gong2025diffullama,ye2025dream}.
While nearly every such adaptation to date has started from a full-attention transformer, AR modeling has meanwhile moved toward \textit{hybrid} architectures that interleave attention with recurrent (RNN) layers~\citep{lieber2024jamba,minimax2025minimax01,qwen2026qwen35}.
Unlike attention, RNN layers are structurally causal and non-trivial to bidirectionalize, which raises a question: can one nevertheless adapt hybrid backbones into performant DLMs?

At first glance, this structural causality seems poorly matched for any-order generation, one of the defining capabilities of DLMs.
Specifically, while an attention causal mask can be easily modified for bidirectionality, the same is not true for the RNN layers that make up most of a hybrid model's layers.
Thus, although each position can, in principle, obtain future context through bidirectional attention layers, the architecture maintains a strong causal bias that conflicts with any-order generation.
Fortunately, this mismatch may be overstated: natural language is produced and read from left to right, and the information useful for predicting a token is concentrated in the tokens immediately preceding it~\citep{khandelwal2018sharp,oconnor2021what,sun2021do}.
In fact, DLMs with full-attention backbones exhibit a global left-to-right decoding trend, even for models like LLaDA that are trained from scratch~\citep{gong2026diffucoder}.
Together, these observations suggest that the RNN's causal bias may, in fact, be complementary to bidirectionalized attention layers (see also~\cref{sec:discussion} for additional discussion).

Motivated by this view, we adapt Qwen3.5 at 0.8B, 2B, 4B, and 9B scales into hybrid-attention DLMs, which we call dQwen3.5.
As suggested above, we bidirectionalize only the attention layers, leave the RNN layers causal, and train all four sizes under a shared adaptation recipe.
Additionally, we adapt Qwen3-1.7B~\citep{yang2025qwen3} as a full-attention control under the same recipe, where notably its trunk (non-embedding, non-head) parameter size closely matches that of Qwen3.5-2B.
With this setup, we then study whether adapted hybrid-attention AR models make for good DLMs: whether they exhibit the any-order and parallel decoding expected of DLMs, and whether such backbones provide a competitive starting point for AR-to-DLM adaptation.

\begin{figure}[t]
\centering
\fitwidth{\includegraphics{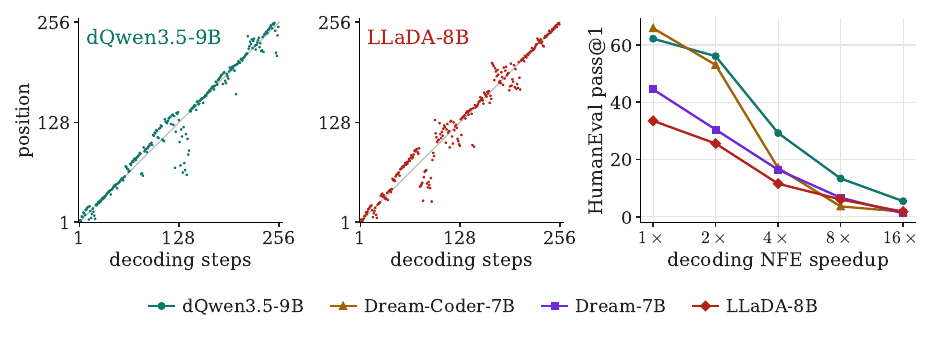}}
\caption{\textbf{Hybrid backbones make strong DLMs.}
Despite causal RNN layers, \dqfam{}-9B supports both any-order and parallel decoding.
\textit{Left, Center:} on HumanEval/27, \dqfam{}-9B and the full-attention, from-scratch LLaDA-8B globally decode left-to-right, but locally decode out-of-order (\cref{sec:eval-order}); each dot marks a position's decode step.
\textit{Right:} at higher NFE speedups, \dqfam{}-9B retains more HumanEval performance than the comparable-trunk DLMs we evaluate.}
\label{fig:intro-teaser}
\end{figure}

Our experiments indicate that hybrid-attention adaptations support both any-order and parallel decoding.
As previewed in~\cref{fig:intro-teaser}, \dqfam{}-9B resembles the full-attention, from-scratch LLaDA-8B: generation trends globally left to right while nearby positions are often decoded out of order.
We quantify this distinction with two decode-order metrics: \emph{local AR-ness} measures left-to-right ordering among adjacent positions (e.g., $k$ and $k+1$), while \emph{global AR-ness} measures it over all pairs (e.g., $i<j$); random order and strict AR decoding score $0.5$ and $1.0$, respectively (\cref{sec:eval-order}).
Across all four \dqfam{} scales, local AR-ness is about $0.64$, within the $0.58$--$0.66$ range of other evaluated DLMs, while global AR-ness is much higher for all models at roughly $0.88$--$0.97$.
For parallel decoding, we measure accuracy against decoding speedup (e.g., $2\times$ decodes two tokens per step; \cref{sec:eval-decode}).
As shown in~\cref{fig:intro-teaser}, \dqfam{}-9B leads the comparable-trunk DLMs we evaluate on HumanEval at every speedup beyond $1\times$.


Beyond their DLM-like decoding behavior, hybrid backbones can also adapt quickly.
Against our trunk-matched full-attention control, \dqfam{}-2B reaches a given training loss in roughly half the tokens.
At this same $\approx$1.4B trunk scale, \dqfam{}-2B after 50B tokens also outperforms CoDA~\citep{chen2025coda} (200B tokens) on 6/7 benchmarks.
At the larger 6.5--7B trunk scale, \dqfam{}-9B after 50B tokens attains the highest score on 4/7 benchmarks against Dream-7B~\citep{ye2025dream} (580B tokens), Dream-Coder-7B~\citep{xie2025dreamcoder} (322B tokens), and LLaDA-8B~\citep{nie2025large} (2.3T tokens).
Interestingly, longer adaptation is not uniformly better: moving from 50B to 100B improves 5/7 benchmarks at 0.8B, but only 1/7 at 9B.


We give the full adaptation recipe in~\cref{sec:recipe} and evaluation in~\cref{sec:eval}.
We further release the base dQwen3.5 models at 0.8B, 2B, 4B, and 9B scales, together with the full-attention control.
With a shared architecture family and adaptation recipe, these releases provide a useful reference for studying hybrid AR-to-DLM adaptation across scale.




\subsection{Related work}
\label{sec:related}

\paragraph{Diffusion language models.}
Discrete diffusion models generate text directly over tokens, rather than in the continuous embedding space of Diffusion-LM~\citep{li2022diffusionlm}.
Notable instances include D3PM~\citep{austin2021structured}, MD4~\citep{shi2024simplified}, MDLM~\citep{sahoo2024simple}, and SEDD~\citep{lou2024discrete}, followed by scaling studies~\citep{nie2024scaling}.
Open releases now range from LLaDA-8B~\citep{nie2025large} and its successor iLLaDA-8B~\citep{nie2026improved} to the 100B-scale LLaDA~2.0~\citep{bie2025llada}, while closed-weight models include Mercury~\citep{inception2025mercury}.
Although these models demonstrate that DLMs can scale to strong language models, doing so from scratch remains substantially more expensive than training comparable AR models, motivating adaptation from pretrained AR checkpoints.


\paragraph{AR-to-DLM adaptation.}
Rather than pretraining from scratch, a DLM can be obtained by adapting a pretrained AR model.
Representative examples include DiffuLLaMA~\citep{gong2025diffullama} (65B tokens), Dream-7B~\citep{ye2025dream} (580B), Dream-Coder-7B~\citep{xie2025dreamcoder} (322B), DiffuCoder~\citep{gong2026diffucoder} (130B), CoDA~\citep{chen2025coda} (200B), Efficient-DLM~\citep{fu2025efficientdlm} (300--500B), and Nemotron-Labs-Diffusion~\citep{fu2026nemotronlabsdiffusion} (300B), compared with roughly 2T--12T tokens for from-scratch DLM pretraining~\citep{nie2025large,nie2026improved}.
Most such adaptations use full-attention backbones.
SDAR~\citep{cheng2025sdar} studies adaptation across the Qwen3 family, but releases only post-trained weights, complicating base-model comparison.
Closest to ours, FLARE~\citep{zhu2026flare} adapts hybrid Qwen3.5 backbones, but does not directly compare hybrid and full-attention backbones under adaptation.




\paragraph{Causal structure in language modeling.}
Natural language itself exhibits a strong left-to-right bias: token predictions depend heavily on nearby prefix context, while distant context contributes more selectively~\citep{khandelwal2018sharp,sun2021do,oconnor2021what}.
Modern hybrid AR models encode this bias architecturally by interleaving attention with causal recurrent layers~\citep{lieber2024jamba,minimax2025minimax01,qwen2026qwen35,yang2025gated}.
DLMs exhibit a similar tendency at inference time: even when decode order is unconstrained, generation progresses broadly left to right, including for models such as LLaDA that were trained from scratch~\citep{gong2026diffucoder}.
Causality can also be imposed explicitly through strategies such as block decoding~\citep{arriola2025block}.
Our work instead asks whether substantial causal structure can remain in the backbone itself while preserving the any-order and parallel decoding behavior of a DLM.

\section{Masked diffusion language models}
\label{sec:background-dlm}
\begin{figure}[t]
\centering
\fitwidth{%
\begin{tikzpicture}[ftok/.style={font=\normalsize, execute at begin node=\strut, inner xsep=0.5pt},
                    fslot/.style={draw=black!25, fill=black!8, rounded corners=1.2pt, minimum width=0.60cm, minimum height=0.34cm, inner sep=0pt},
                    fframe/.style={draw=epaccent!65, dash pattern=on 2pt off 1.6pt, rounded corners=2.5pt, line width=0.7pt}]
  \node[figmeta, font=\normalsize] at (5.205,0.60) {full-canvas decoding};
  \node[figmeta, font=\normalsize] at (16.165,0.60) {block decoding (block size 4)};
  \draw[figbox] (-0.060,-0.350) rectangle (10.470,0.350);
  \node[ftok, text=epmuted] at (0.440,0.00) {The};
  \node[ftok, text=epmuted] at (1.330,0.00) {quick};
  \node[fslot] at (2.510,0.00) {};
  \node[fslot] at (3.560,0.00) {};
  \node[fslot] at (4.610,0.00) {};
  \node[fslot] at (5.660,0.00) {};
  \node[fslot] at (6.710,0.00) {};
  \node[fslot] at (7.760,0.00) {};
  \node[fslot] at (8.810,0.00) {};
  \node[fslot] at (9.860,0.00) {};
  \draw[figbox] (10.900,-0.350) rectangle (21.430,0.350);
  \node[ftok, text=epmuted] at (11.400,0.00) {The};
  \node[ftok, text=epmuted] at (12.290,0.00) {quick};
  \node[fslot] at (13.470,0.00) {};
  \node[fslot] at (14.520,0.00) {};
  \node[fslot] at (15.570,0.00) {};
  \node[fslot] at (16.620,0.00) {};
  \node[fslot] at (17.670,0.00) {};
  \node[fslot] at (18.720,0.00) {};
  \node[fslot] at (19.770,0.00) {};
  \node[fslot] at (20.820,0.00) {};
  \draw[fframe] (12.885,-0.280) rectangle (17.205,0.280);
  \draw[figbox] (-0.060,-1.300) rectangle (10.470,-0.600);
  \node[ftok, text=epmuted] at (0.440,-0.95) {The};
  \node[ftok, text=epmuted] at (1.330,-0.95) {quick};
  \node[fslot] at (2.510,-0.95) {};
  \node[ftok] at (3.560,-0.95) {\textbf{fox}};
  \node[fslot] at (4.610,-0.95) {};
  \node[fslot] at (5.660,-0.95) {};
  \node[fslot] at (6.710,-0.95) {};
  \node[ftok] at (7.760,-0.95) {\textbf{lazy}};
  \node[fslot] at (8.810,-0.95) {};
  \node[fslot] at (9.860,-0.95) {};
  \draw[figbox] (10.900,-1.300) rectangle (21.430,-0.600);
  \node[ftok, text=epmuted] at (11.400,-0.95) {The};
  \node[ftok, text=epmuted] at (12.290,-0.95) {quick};
  \node[ftok] at (13.530,-0.95) {\textbf{brown}};
  \node[fslot] at (14.520,-0.95) {};
  \node[ftok] at (15.570,-0.95) {\textbf{jumps}};
  \node[fslot] at (16.620,-0.95) {};
  \node[fslot] at (17.670,-0.95) {};
  \node[fslot] at (18.720,-0.95) {};
  \node[fslot] at (19.770,-0.95) {};
  \node[fslot] at (20.820,-0.95) {};
  \draw[fframe] (12.885,-1.230) rectangle (17.205,-0.670);
  \draw[figbox] (-0.060,-2.250) rectangle (10.470,-1.550);
  \node[ftok, text=epmuted] at (0.440,-1.90) {The};
  \node[ftok, text=epmuted] at (1.330,-1.90) {quick};
  \node[ftok] at (2.570,-1.90) {\textbf{brown}};
  \node[ftok] at (3.560,-1.90) {fox};
  \node[fslot] at (4.610,-1.90) {};
  \node[fslot] at (5.660,-1.90) {};
  \node[fslot] at (6.710,-1.90) {};
  \node[ftok] at (7.760,-1.90) {lazy};
  \node[ftok] at (8.810,-1.90) {\textbf{dog}};
  \node[fslot] at (9.860,-1.90) {};
  \draw[figbox] (10.900,-2.250) rectangle (21.430,-1.550);
  \node[ftok, text=epmuted] at (11.400,-1.90) {The};
  \node[ftok, text=epmuted] at (12.290,-1.90) {quick};
  \node[ftok] at (13.530,-1.90) {brown};
  \node[ftok] at (14.520,-1.90) {\textbf{fox}};
  \node[ftok] at (15.570,-1.90) {jumps};
  \node[ftok] at (16.620,-1.90) {\textbf{over}};
  \node[fslot] at (17.670,-1.90) {};
  \node[fslot] at (18.720,-1.90) {};
  \node[fslot] at (19.770,-1.90) {};
  \node[fslot] at (20.820,-1.90) {};
  \draw[fframe] (17.145,-2.180) rectangle (21.345,-1.620);
  \draw[figbox] (-0.060,-3.200) rectangle (10.470,-2.500);
  \node[ftok, text=epmuted] at (0.440,-2.85) {The};
  \node[ftok, text=epmuted] at (1.330,-2.85) {quick};
  \node[ftok] at (2.570,-2.85) {brown};
  \node[ftok] at (3.560,-2.85) {fox};
  \node[ftok] at (4.610,-2.85) {\textbf{jumps}};
  \node[fslot] at (5.660,-2.85) {};
  \node[ftok] at (6.710,-2.85) {\textbf{the}};
  \node[ftok] at (7.760,-2.85) {lazy};
  \node[ftok] at (8.810,-2.85) {dog};
  \node[fslot] at (9.860,-2.85) {};
  \draw[figbox] (10.900,-3.200) rectangle (21.430,-2.500);
  \node[ftok, text=epmuted] at (11.400,-2.85) {The};
  \node[ftok, text=epmuted] at (12.290,-2.85) {quick};
  \node[ftok] at (13.530,-2.85) {brown};
  \node[ftok] at (14.520,-2.85) {fox};
  \node[ftok] at (15.570,-2.85) {jumps};
  \node[ftok] at (16.620,-2.85) {over};
  \node[fslot] at (17.670,-2.85) {};
  \node[ftok] at (18.720,-2.85) {\textbf{lazy}};
  \node[fslot] at (19.770,-2.85) {};
  \node[ftok] at (20.820,-2.85) {\textbf{.}};
  \draw[fframe] (17.145,-3.130) rectangle (21.345,-2.570);
  \draw[figbox] (-0.060,-4.150) rectangle (10.470,-3.450);
  \node[ftok, text=epmuted] at (0.440,-3.80) {The};
  \node[ftok, text=epmuted] at (1.330,-3.80) {quick};
  \node[ftok] at (2.570,-3.80) {brown};
  \node[ftok] at (3.560,-3.80) {fox};
  \node[ftok] at (4.610,-3.80) {jumps};
  \node[ftok] at (5.660,-3.80) {\textbf{over}};
  \node[ftok] at (6.710,-3.80) {the};
  \node[ftok] at (7.760,-3.80) {lazy};
  \node[ftok] at (8.810,-3.80) {dog};
  \node[ftok] at (9.860,-3.80) {\textbf{.}};
  \draw[figbox] (10.900,-4.150) rectangle (21.430,-3.450);
  \node[ftok, text=epmuted] at (11.400,-3.80) {The};
  \node[ftok, text=epmuted] at (12.290,-3.80) {quick};
  \node[ftok] at (13.530,-3.80) {brown};
  \node[ftok] at (14.520,-3.80) {fox};
  \node[ftok] at (15.570,-3.80) {jumps};
  \node[ftok] at (16.620,-3.80) {over};
  \node[ftok] at (17.670,-3.80) {\textbf{the}};
  \node[ftok] at (18.720,-3.80) {lazy};
  \node[ftok] at (19.770,-3.80) {\textbf{dog}};
  \node[ftok] at (20.820,-3.80) {.};
  \draw[-{Stealth[length=4pt,width=3.5pt]}, black!35, line width=0.7pt] (-0.40,0.00) -- (-0.40,-3.80);
  \node[figmeta, font=\normalsize, rotate=90] at (-0.62,-1.90) {decoding steps};
\end{tikzpicture}}

\caption{\textbf{Full-canvas and block decoding.}
Gray slots denote masks, while the prompt (``The quick'') remains fixed.
Both schemes begin from the same masked canvas and decode 2 tokens per step.
Full-canvas decoding (left) may unmask anywhere, whereas block decoding (right) restricts unmasking to the active block (dashed outline), advancing to the next block once it is complete.
}

\label{fig:block-decoding}
\end{figure}
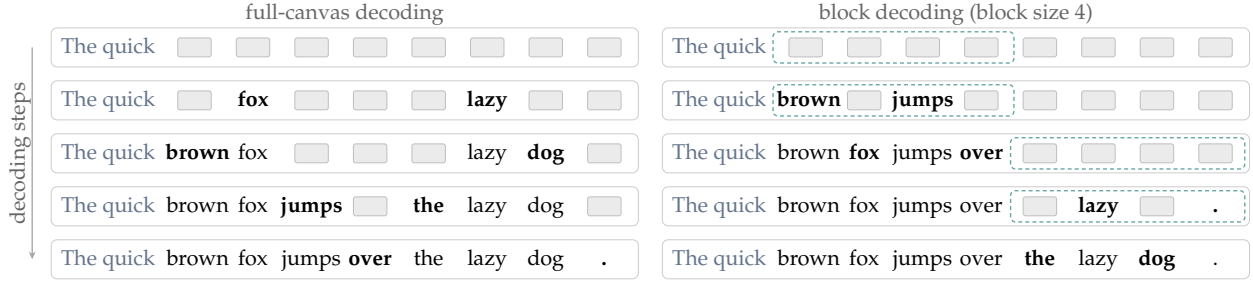

A masked diffusion language model (DLM) generates text by iteratively decoding (unmasking) masked positions rather than generating strictly from left to right \citep{austin2021structured,shi2024simplified,sahoo2024simple}.
During training, positions are randomly masked at varying rates and the model predicts the conditional marginal distribution (equivalently, the logits) corresponding to the original tokens at the masked positions, allowing each prediction to use context from both its left and right.
At generation time, prompt tokens remain fixed while the masked positions form a \emph{canvas}; at each step, the model predicts token logits for the masked positions and a subset is unmasked (sampled from the logits) until the canvas is fully decoded.
Which positions to decode are commonly selected by confidence (top-1 token probability), either through top-$k$ or threshold-based schemes~\citep{nie2025large,ye2025dream}.

In~\cref{fig:block-decoding}, an 8-position canvas unmasks 2 tokens per step and is therefore fully decoded in 4 steps.
\emph{Full-canvas decoding} may decode positions anywhere on the canvas, whereas \emph{block decoding} partitions it into fixed-size blocks---of size 4 in the figure---and completes each block before advancing, yielding a semi-autoregressive trajectory that is left-to-right across blocks but any-order within the active block \citep{arriola2025block}.
We study both in~\cref{sec:eval-decode} and~\cref{app:block}, and give the formal diffusion process and training objective in~\cref{app:objective}.

\subsection{Recurrent and hybrid language-model backbones}
\label{sec:background-arch}
\begin{figure}[t]
\centering
%
\begin{minipage}[c]{0.1985\linewidth}
\centering
\begin{tikzpicture}[
  zfa/.style={draw=none, fill=epaccent!92!black, rounded corners=2pt},
  zgdn/.style={draw=epaccent!30, line width=0.5pt, fill=epaccent!8, rounded corners=2pt},
  rowlab/.style={font=\scriptsize, anchor=center, inner sep=0pt, execute at begin node=\strut},
  nodesc/.style={rowlab, yshift=-0.70pt},
]
  \path (0.45, -0.328) rectangle (3.70, 2.856);
  \draw[draw=black!25, rounded corners=4pt, line width=0.6pt] (0.45, -0.159) rectangle (3.70, 2.856);
  \node[font=\scriptsize, anchor=west, inner xsep=3pt, inner ysep=0pt,
        fill=white, text=black!60, execute at begin node=\strut] at (0.72, 2.856) {\textsc{Hybrid Block}};
  \draw[zgdn] (0.7, 0.000) rectangle (3.45, 0.514);
  \node[nodesc, text=black!75] at (2.075, 0.257) {Gated DeltaNet};
  \draw[zgdn] (0.7, 0.683) rectangle (3.45, 1.197);
  \node[nodesc, text=black!75] at (2.075, 0.940) {Gated DeltaNet};
  \draw[zgdn] (0.7, 1.366) rectangle (3.45, 1.880);
  \node[nodesc, text=black!75] at (2.075, 1.623) {Gated DeltaNet};
  \draw[zfa] (0.7, 2.049) rectangle (3.45, 2.563);
  \node[rowlab, text=white] at (2.075, 2.306) {Attention Layer};
\end{tikzpicture}
\end{minipage}\hspace{0.0664\linewidth}%
\begin{minipage}[c]{0.6640\linewidth}
\centering
\footnotesize
\begin{tabular}{lcrrrr}
\toprule
backbone & GDN + attention layers & trunk & embed & head & total \\
\midrule
Qwen3.5-0.8B & $18 + 6$ & 0.50B & 0.25B & tied & 0.75B \\
Qwen3.5-2B & $18 + 6$ & 1.37B & 0.51B & tied & 1.88B \\
Qwen3.5-4B & $24 + 8$ & 3.57B & 0.64B & tied & 4.21B \\
Qwen3.5-9B & $24 + 8$ & 6.92B & 1.02B & 1.02B & 8.95B \\
\midrule
Qwen3-1.7B & $0 + 28$ & 1.41B & 0.31B & tied & 1.72B \\
\bottomrule
\end{tabular}
\end{minipage}

\caption{\textbf{Bidirectionalizing a hybrid model.}
Each hybrid block applies three GDNs before a single attention layer;
we bidrectionalize only the attention and leave the GDNs causal.
The table reports layer and parameter counts across scales, including the Qwen3-1.7B full-attention control.
}

\label{fig:hybrid-block}
\end{figure}
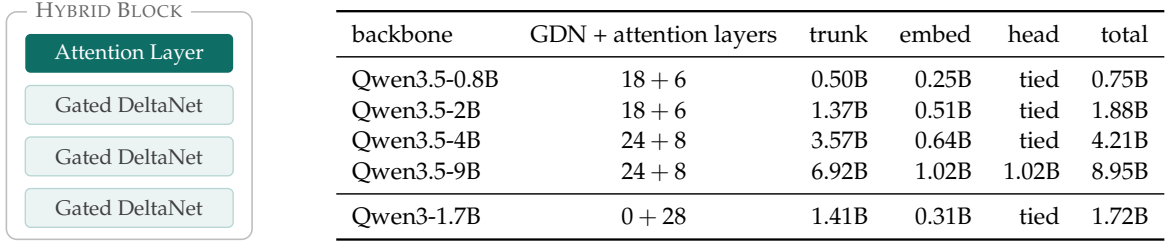

A growing class of language models interleaves self-attention with recurrent neural network (RNN) layers, often making the recurrent layers the majority of the architecture stack \citep{lieber2024jamba,minimax2025minimax01,yang2025gated,qwen2026qwen35}.
Gated DeltaNets (GDNs) are one such modern RNN: they scan the sequence left to right while compressing the prefix into a fixed-size recurrent state \citep{yang2025gated}.
Qwen3.5 uses GDNs as its main sequence-processing layer, with three GDN layers for every standard attention layer \citep{qwen2026qwen35} (see~\cref{fig:hybrid-block}).
Consequently, causality in a GDN is built into its recurrence rather than imposed by an attention mask, and so bidirectionalization would require non-trivial architectural modifications.



\section{Adaptation recipe}
\label{sec:recipe}

Our adaptation follows standard AR-to-DLM practice wherever possible; the main difference is that only Qwen3.5's attention layers are made bidirectional, while its recurrent layers remain causal.
We make no optimality claims about individual recipe choices and report the exact configuration used in our experiments.

\subsection{Architecture modifications}
\label{sec:recipe-surgery}

We adapt Qwen3.5-0.8B, Qwen3.5-2B, Qwen3.5-4B, and Qwen3.5-9B.
These checkpoints include vision components and an auxiliary multi-token-prediction head; we discard both and adapt only the language-model backbone, which is also why our parameter counts differ from those reported on the model cards.
The language-model backbone consists of stacks of hybrid blocks (\cref{fig:hybrid-block}); we modify only their attention layers and otherwise leave the stack structurally unchanged.
For comparison, we also adapt Qwen3-1.7B as a full-attention control, whose 1.41B-parameter trunk closely matches the 1.37B-parameter trunk of Qwen3.5-2B.

\paragraph{Bidirectionalization.}
Across all four Qwen3.5 sizes, attention makes up 25\% of the sequence-processing stack (6/24 layers for the two smaller models and 8/32 for the two larger ones); we make these layers bidirectional by disabling their causal masks.
The remaining Gated DeltaNet layers are causal by construction (\cref{sec:background-arch}) and are left untouched.
Thus, only a minority of the stack becomes bidirectional, while the recurrent majority retains the causal structure studied in this work.
For the full-attention control, every sequence-processing layer is attention and is therefore bidirectionalized.
The remaining modifications follow standard AR-to-DLM adaptation practice.

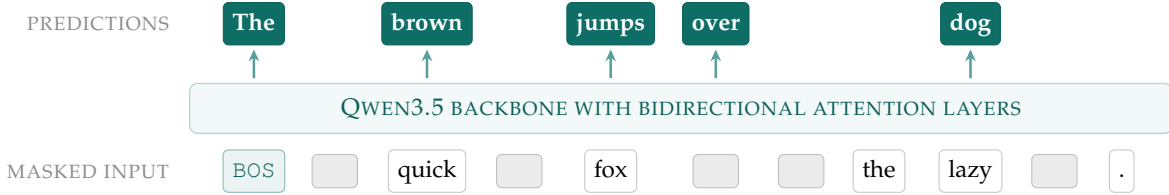
\begin{figure}[t]
\centering
\begin{tikzpicture}[
  chip/.style={draw=black!22, fill=white, rounded corners=2pt, inner xsep=3.5pt, inner ysep=2.2pt, font=\footnotesize, execute at begin node=\strut},
  chipbos/.style={chip, draw=epaccent!50, fill=epaccent!8, font=\ttfamily\footnotesize, text=epaccent!80!black},
  chippay/.style={chip, draw=epaccent!85!black, fill=epaccent!92!black, text=white, font=\footnotesize\bfseries},
  mslot/.style={draw=black!25, fill=black!8, rounded corners=2pt, minimum width=0.60cm, minimum height=0.44cm, inner sep=0pt},
  hpay/.style={-{Stealth[length=4.5pt,width=4pt]}, line width=0.95pt, color=epaccent!70, shorten >=2pt, shorten <=2pt},
]
  \node[font=\footnotesize\scshape, text=black!45, anchor=east] at (-0.55,0) {masked input};
  \node[chipbos] at (0.43,0) {BOS};
  \node[mslot] at (1.50,0) {};
  \node[chip] at (2.72,0) {quick};
  \node[mslot] at (3.94,0) {};
  \node[chip] at (5.15,0) {fox};
  \node[mslot] at (6.54,0) {};
  \node[mslot] at (7.67,0) {};
  \node[chip] at (8.70,0) {the};
  \node[chip] at (9.91,0) {lazy};
  \node[mslot] at (11.02,0) {};
  \node[chip] at (11.91,0) {.};
  \draw[figbox, fill=epaccent!5, draw=epaccent!30] (-0.42,0.50) rectangle (12.58,1.16);
  \node[font=\footnotesize\scshape, text=epaccent!80!black] at (6.08,0.83) {Qwen3.5 backbone with bidirectional attention layers};
  \node[font=\footnotesize\scshape, text=black!45, anchor=east] at (-0.55,1.95) {predictions};
  \node[chippay] at (0.43,1.95) {The};
  \draw[hpay] (0.43,1.16) -- (0.43,1.68);
  \node[chippay] at (2.72,1.95) {brown};
  \draw[hpay] (2.72,1.16) -- (2.72,1.68);
  \node[chippay] at (5.15,1.95) {jumps};
  \draw[hpay] (5.15,1.16) -- (5.15,1.68);
  \node[chippay] at (6.54,1.95) {over};
  \draw[hpay] (6.54,1.16) -- (6.54,1.68);
  \node[chippay] at (9.91,1.95) {dog};
  \draw[hpay] (9.91,1.16) -- (9.91,1.68);
\end{tikzpicture}

\caption{\textbf{Token shifting preserves the AR readout alignment.}
The hidden state at position $k$ predicts the token at position $k+1$, matching the readout used during AR pretraining while allowing bidirectional attention to incorporate future context.
This alignment is known to improve AR-to-DLM adaptation~\citep{gong2025diffullama,ye2025dream}.}

\label{fig:decoding-pass}
\end{figure}

\paragraph{Token shifting.}
We prepend a \texttt{BOS} token and shift the readout by one position, so that the hidden state at position $k$ predicts the token at position $k+1$ when that target position is masked (\cref{fig:decoding-pass}).
This preserves the next-token alignment of the pretrained AR model while allowing the input itself to contain masked tokens.
Token shifting is standard across the AR-to-DLM literature and is known to help adaptation \citep{gong2025diffullama,ye2025dream,xie2025dreamcoder}.

\paragraph{Special tokens.}
We additionally need three special tokens for diffusion language modeling with token shifting: a mask token for the diffusion objective, a padding token the model emits to halt generation, and a leading BOS for the shifted readout.
We repurpose unused ids rather than resizing the embedding table: mask is Qwen3.5's \texttt{<|fim\_middle|>}, padding its \texttt{<|endoftext|>}, and BOS its \texttt{<tts\_text\_bos>}.

\subsection{Training data and configuration}
\label{sec:recipe-data}

We give an overview here; the complete training configuration is in \cref{app:asrun}, with our learning-rate and mixture explorations summarized in \cref{app:devnotes}.

\paragraph{Mixture.}
We train on a fixed mixture of 14 subsets drawn from NVIDIA's Nemotron pretraining releases\footnote{\url{https://huggingface.co/collections/nvidia/nemotron-pre-training-datasets}}
\citep{su2024nemotroncc,nvidia2025nemotronnano2,nvidia2025nemotron3nano,karimimahabadi2025nemotronccmath}: 50\% code, 35\% general text, and 15\% mathematics by token share; \cref{app:asrun} lists the subsets and their weights.
Sequences are packed to 4096 tokens, and a small fraction is randomly truncated so that not every example fills the context.
Before applying diffusion masking, each 512-sequence batch contains about 1.99M content (non-mask, non-pad) tokens on average after truncation and padding.

\paragraph{Training.}
We train a time-reweighted form of the standard masked diffusion objective \citep{austin2021structured,shi2024simplified,sahoo2024simple}, following \citet{shi2025demystifying}, defined in \cref{app:objective}.
Across all sizes, we use the same optimizer and warmup--stable--decay schedule, with only the learning rate varying by model size ($3, 2, 2, 1 \times 10^{-5}$ for 0.8B/2B/4B/9B).
The global batch is 512 sequences $\times$ 4096 tokens.
For each size, we retain checkpoints at two budgets: 25k steps (1k\,/\,19k\,/\,5k, 50B content tokens) and 50k steps (1k\,/\,44k\,/\,5k, 100B content tokens).
These two budgets let us study how quickly useful diffusion capabilities emerge and whether further adaptation continues to help.
We checkpoint every 1{,}000 steps and release the weight average of the last five checkpoints: steps 21{,}000--25{,}000 at the shorter budget and 46{,}000--50{,}000 at the longer.

\subsection{Training dynamics}
\label{sec:recipe-dynamics}

The training curves are smooth across sizes, but loss should be interpreted carefully.
In particular, the 4B and 9B loss curves in \cref{fig:telemetry} nearly overlap despite separating clearly on downstream benchmarks (\cref{sec:eval}), so we use training loss primarily to compare adaptation speed in the closely matched hybrid/control setting rather than as a universal measure of capability.
All runs begin from the same mixture under the same seed, although wall-clock limits require periodic resumes that reshuffle the data and cause their sample streams to diverge.
We therefore evaluate fixed 50B- and 100B-token checkpoints rather than train to convergence.
Loss is still declining at the end of every run, but prior work has found that longer AR-to-DLM adaptation can nevertheless degrade downstream performance \citep{gong2026diffucoder}.

\section{Evaluation}
\label{sec:eval}

\begin{table}[t]
\centering
\caption{\textbf{Evaluated models.} Backbones are hybrid (H) or full (F) attention.}
\label{tab:model-registry}
\footnotesize
\begin{tabular}{lllrrrr}
\toprule
 & & & \multicolumn{4}{c}{parameters} \\
\cmidrule(lr){4-7}
model & attn & checkpoint & trunk & embed & head & total \\
\midrule
\multicolumn{7}{l}{\emph{ours}} \\
\dqfam{}-0.8B & H & \texttt{\scriptsize UT-IFML/dQwen3.5-0.8B-Base} & 0.50B & 0.25B & tied & 0.75B \\
\dqfam{}-2B & H & \texttt{\scriptsize UT-IFML/dQwen3.5-2B-Base} & 1.37B & 0.51B & tied & 1.88B \\
\dqfam{}-4B & H & \texttt{\scriptsize UT-IFML/dQwen3.5-4B-Base} & 3.57B & 0.64B & tied & 4.21B \\
\dqfam{}-9B & H & \texttt{\scriptsize UT-IFML/dQwen3.5-9B-Base} & 6.92B & 1.02B & 1.02B & 8.95B \\
\dqctrl{}-1.7B & F & \texttt{\scriptsize UT-IFML/dQwen3-1.7B-Base} & 1.41B & 0.31B & tied & 1.72B \\
\midrule
\multicolumn{7}{l}{\emph{diffusion comparators}} \\
CoDA & F & \texttt{\scriptsize Salesforce/CoDA-v0-Base} & 1.41B & 0.31B & 0.31B & 2.03B \\
LLaDA-8B & F & \texttt{\scriptsize GSAI-ML/LLaDA-8B-Base} & 6.98B & 0.52B & 0.52B & 8.02B \\
Dream-7B & F & \texttt{\scriptsize Dream-org/Dream-v0-Base-7B} & 6.53B & 0.54B & 0.54B & 7.62B \\
Dream-Coder-7B & F & \texttt{\scriptsize Dream-org/Dream-Coder-v0-Base-7B} & 6.53B & 0.54B & 0.54B & 7.62B \\
\midrule
\multicolumn{7}{l}{\emph{autoregressive parents}} \\
Qwen3.5-0.8B & H & \texttt{\scriptsize Qwen/Qwen3.5-0.8B} & 0.50B & 0.25B & tied & 0.75B \\
Qwen3.5-2B & H & \texttt{\scriptsize Qwen/Qwen3.5-2B} & 1.37B & 0.51B & tied & 1.88B \\
Qwen3.5-4B & H & \texttt{\scriptsize Qwen/Qwen3.5-4B} & 3.57B & 0.64B & tied & 4.21B \\
Qwen3.5-9B & H & \texttt{\scriptsize Qwen/Qwen3.5-9B} & 6.92B & 1.02B & 1.02B & 8.95B \\
Qwen3-1.7B & F & \texttt{\scriptsize Qwen/Qwen3-1.7B} & 1.41B & 0.31B & tied & 1.72B \\
Qwen2.5-7B & F & \texttt{\scriptsize Qwen/Qwen2.5-7B} & 6.53B & 0.54B & 0.54B & 7.62B \\
Qwen2.5-Coder-7B & F & \texttt{\scriptsize Qwen/Qwen2.5-Coder-7B} & 6.53B & 0.54B & 0.54B & 7.62B \\
\bottomrule
\end{tabular}
\end{table}


We evaluate whether a mostly causal hybrid backbone can nevertheless become an effective DLM.
We focus on two questions: whether hybrid backbones provide efficient starting points for AR-to-DLM adaptation, and whether the resulting models retain the parallel and any-order decoding behavior that defines DLMs.
We evaluate downstream capability on standard knowledge, mathematics, and coding benchmarks: MMLU~\citep{hendrycks2021mmlu}, GSM8K~\citep{cobbe2021gsm8k}, MATH500~\citep{hendrycks2021math,lightman2023verify}, HumanEval~\citep{chen2021humaneval}, and MBPP~\citep{austin2021program}, along with the EvalPlus variants of the last two~\citep{liu2023evalplus}; complete evaluation details are in~\cref{app:evalx}.

\paragraph{Models.}
\Cref{tab:model-registry} lists all evaluated models and their parameter decompositions.
We compare base DLM checkpoints throughout, including our own models before post-training.
Post-training can substantially change downstream performance and is highly recipe-sensitive, so we treat it as a separate problem from AR-to-DLM adaptation.
We use trunk (non-embedding, non-head) parameters as a proxy for capacity, yielding comparison groups at the ${\approx}1.4$B and $6.5$--$7$B scales.
Dream-7B, Dream-Coder-7B, and CoDA are themselves adapted from Qwen2.5-7B~\citep{qwen2024qwen25}, Qwen2.5-Coder-7B~\citep{hui2024qwen25coder}, and Qwen3-1.7B~\citep{yang2025qwen3}, respectively; the last is also the AR model used for our control.
CoDA unties its embedding and output head, giving 2.03B total parameters despite a 1.41B trunk.

\paragraph{Decoding.}
We evaluate all DLMs under a common decoding scheme rather than their released inference recipes.
By default, we use a generation canvas of size 1024 and a block size of 32 for block decoding.
For parallel decoding, we use either a fixed decoding rate or confidence thresholding.
For confidence thresholding, we unmask all positions above the threshold, while always unmasking at least enough highest-confidence positions to satisfy the linear schedule $\alpha_t=1-t$.
Otherwise, we greedily unmask only the single most confident position at each step.

\subsection{Hybrid backbones adapt faster}
\label{sec:eval-hybrid}

\begin{figure}[t]
\centering
\fitwidth{\includegraphics{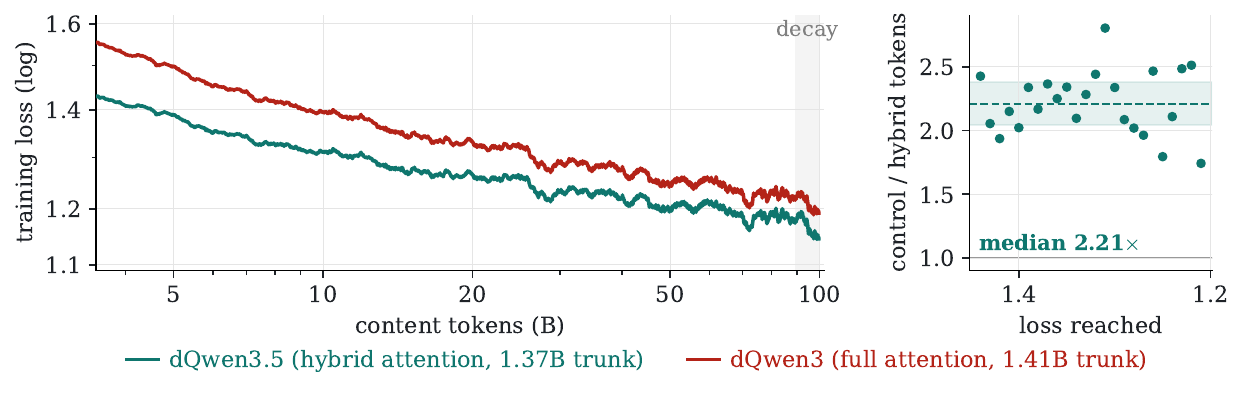}}
\caption{\textbf{Hybrid attention reaches the same loss in about half the tokens.} \dqfam{}-2B (1.37B trunk) against \dqctrl{}-1.7B (1.41B trunk).
\emph{Left:} training loss after 2B warmup tokens.
\emph{Right:} at each stable-phase loss level, the ratio of tokens each run needed to first reach it.
}
\label{fig:convergence-vs-control}
\end{figure}

We first ask whether a hybrid backbone provides a more efficient starting point for AR-to-DLM adaptation.
We compare \dqfam{}-2B against \dqctrl{}-1.7B under the same adaptation recipe, with closely matched trunks of 1.37B and 1.41B parameters, respectively.
Importantly, the architectural intervention is very different: only 6/24 sequence-processing layers are bidirectionalized in \dqfam{}-2B, compared with all 28 layers in the full-attention control.
This pair therefore gives our closest test of whether retaining a causal recurrent majority can make adaptation more efficient.

Against this control, the hybrid reaches the diffusion objective substantially faster (\cref{fig:convergence-vs-control}).
Across stable-phase loss levels reached by both runs, the control requires a median $2.21\times$ as many content tokens to reach the same loss.
We interpret this as faster adaptation to the diffusion objective, rather than as evidence that training loss directly measures downstream capability.
The comparison is not exact because the tokenizers differ: on our mixture, \dqctrl{} averages 4.18 bytes/token versus 4.05 for \dqfam{}, which complicates a direct comparison of per-token losses.

\begin{figure}[t]
\centering
\fitwidth{\includegraphics{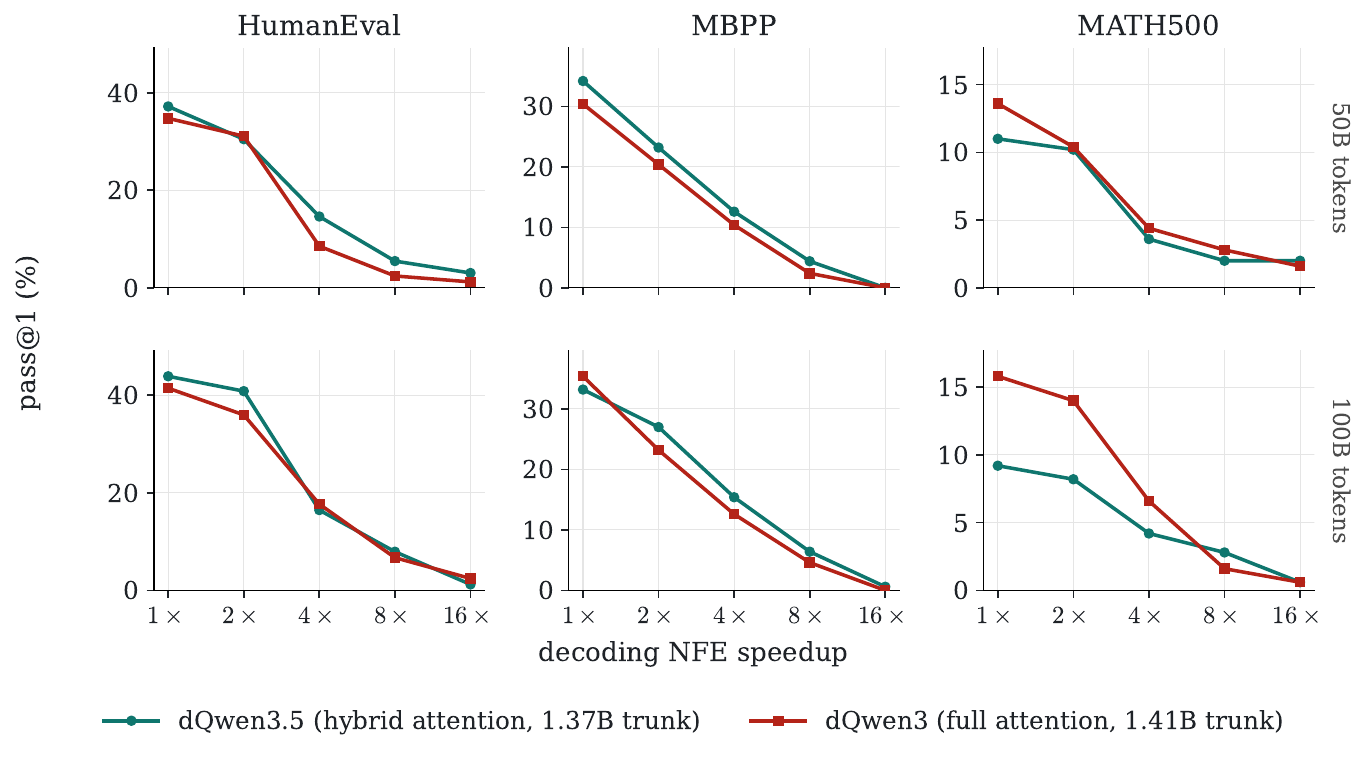}}
\caption{\textbf{Hybrid adapts faster and can parallelize.}
Accuracy against decoding speedup, where $1\times$ is one token per step and 1024 NFEs.
Notably, Qwen3.5-2B (AR, hybrid) actually underperforms Qwen3-1.7B (AR, full) on every benchmark (\cref{tab:budget-trade}).
}
\label{fig:decode-budget}
\end{figure}

The downstream results provide a second positive signal.
Before adaptation, Qwen3.5-2B underperforms Qwen3-1.7B on all 7 benchmarks we evaluate, yet after 50B adaptation tokens \dqfam{}-2B overtakes the full-attention control on 3/7, including HumanEval, MBPP, and MBPP+ (\cref{tab:budget-trade}).
The hybrid also often remains ahead on code under parallel decoding (\cref{fig:decode-budget}).
It remains weaker on mathematics, and these downstream differences cannot be attributed solely to the backbone because the AR parents already differ substantially and our code-heavy adaptation mixture also affects performance (\cref{app:training}).
Nevertheless, reversing several initially unfavorable comparisons provides further evidence that the hybrid is an effective starting point for adaptation.




\subsection{50B tokens is enough for adaptation}
\label{sec:eval-budget}
\begin{table}[t]
\centering
\caption{\textbf{Too much adaptation can hurt.} Each model at its 50B- and 100B-token checkpoints, per~\cref{sec:recipe}.
We bold the higher of each 50B/100B pair.
}
\label{tab:budget-trade}
\footnotesize
\begin{tabular}{llrrrrrrr}
\toprule
model & tokens & MMLU & GSM8K & MATH500 & HEval & HEval+ & MBPP & MBPP+ \\
\midrule
Qwen3.5-0.8B & --- & 50.31 & 34.95 & 17.00 & 22.56 & 20.73 & 17.40 & 24.07 \\
\dqfam{}-0.8B & 50B & \textbf{32.62} & \textbf{8.49} & 5.40 & 28.05 & 25.61 & 22.60 & 29.10 \\
               & 100B & 30.24 & 8.11 & \textbf{6.20} & \textbf{35.37} & \textbf{31.71} & \textbf{23.20} & \textbf{32.28} \\
\addlinespace
Qwen3.5-2B & --- & 57.23 & 58.98 & 30.80 & 37.20 & 34.15 & 23.40 & 31.75 \\
\dqfam{}-2B & 50B & \textbf{45.50} & \textbf{20.55} & \textbf{11.00} & 37.20 & 32.32 & \textbf{34.20} & 40.48 \\
               & 100B & 41.55 & 19.79 & 9.20 & \textbf{43.90} & \textbf{38.41} & 33.20 & \textbf{43.12} \\
\addlinespace
Qwen3.5-4B & --- & 69.82 & 85.29 & 49.60 & 59.15 & 53.66 & 40.20 & 48.41 \\
\dqfam{}-4B & 50B & \textbf{62.42} & \textbf{57.77} & 25.00 & 54.88 & 51.83 & 44.60 & \textbf{51.32} \\
               & 100B & 57.95 & 53.45 & \textbf{25.60} & \textbf{60.37} & \textbf{57.32} & 44.60 & 50.00 \\
\addlinespace
Qwen3.5-9B & --- & 69.85 & 87.34 & 48.80 & 68.90 & 62.80 & 55.00 & 63.76 \\
\dqfam{}-9B & 50B & \textbf{74.57} & \textbf{74.98} & \textbf{37.40} & \textbf{64.02} & \textbf{59.15} & 50.20 & \textbf{58.20} \\
               & 100B & 72.78 & 69.98 & 36.80 & 62.20 & 58.54 & \textbf{52.00} & 56.61 \\
\addlinespace
Qwen3-1.7B & --- & 60.30 & 74.75 & 48.40 & 40.24 & 37.80 & 43.80 & 46.03 \\
\dqctrl{}-1.7B & 50B & \textbf{51.30} & \textbf{42.23} & 13.60 & 34.76 & 33.54 & 30.40 & 38.89 \\
               & 100B & 48.12 & 39.35 & \textbf{15.80} & \textbf{41.46} & \textbf{37.20} & \textbf{35.40} & \textbf{46.83} \\
\bottomrule
\end{tabular}
\end{table}

\begin{figure}[tp]
\centering
\fitwidth{\includegraphics{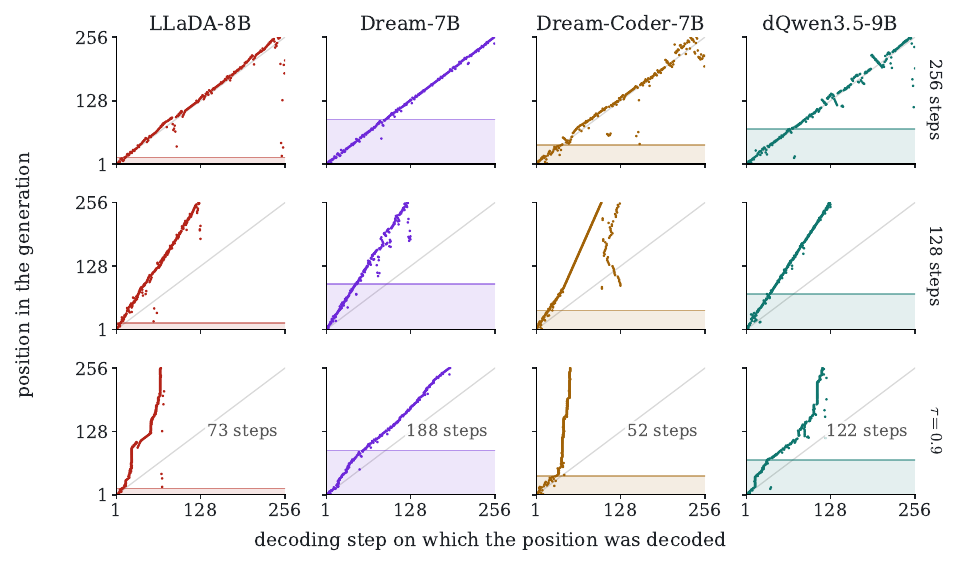}}
\\[8pt] \footnotesize
\setlength{\tabcolsep}{6pt}
\begin{tabular}{lccccccc}
\toprule
 & \multicolumn{2}{c}{full-canvas 256} & \multicolumn{2}{c}{full-canvas 128}
 & \multicolumn{3}{c}{full-canvas $\tau = 0.9$} \\
\cmidrule(lr){2-3}\cmidrule(lr){4-5}\cmidrule(lr){6-8}
model & $\mathrm{ARL}$ & $\mathrm{ARG}$ & $\mathrm{ARL}$ & $\mathrm{ARG}$ & $\mathrm{ARL}$ & $\mathrm{ARG}$ & NFEs \\
\midrule
LLaDA-8B & 0.631 & 0.932 & 0.617 & 0.924 & 0.630 & 0.945 & $83 \pm 38$ \\
Dream-7B & 0.638 & 0.884 & 0.619 & 0.830 & 0.659 & 0.891 & $139 \pm 40$ \\
Dream-Coder-7B & 0.652 & 0.937 & 0.607 & 0.934 & 0.662 & 0.944 & $136 \pm 55$ \\
\dqfam{}-9B & 0.636 & 0.967 & 0.651 & 0.970 & 0.655 & 0.975 & $129 \pm 34$ \\
\midrule
AR decoder & 1.000 & 1.000 & 1.000 & 1.000 & 1.000 & 1.000 & $256$ \\
\bottomrule
\end{tabular}

\normalsize
\caption{\textbf{Decode orders for the $6.5$--$7$B trunks on \texttt{HumanEval/58}.}
We use canvas size 256 because extracted solutions (shaded) tend to be short.
\textit{Top:} decode position vs.\ step number; each dot marks when and where a position is unmasked.
\textit{Bottom:} $\mathrm{ARL}$ and $\mathrm{ARG}$ over all 164 HumanEval problems.
All DLMs decode globally left-to-right but remain locally flexible, including the mostly causal \dqfam{}-9B and the from-scratch LLaDA-8B.
See~\cref{app:decodebehav} for additional results.}

\label{fig:decode-order}
\end{figure}

A central goal of AR-to-DLM adaptation is to obtain useful diffusion capability with substantially less training than pretraining from scratch.
We therefore evaluate every \dqfam{} size after both 50B and 100B adaptation tokens (\cref{tab:budget-trade}).
Across scales, the 50B checkpoints are already strong, and extending training to 100B tokens does not uniformly improve downstream performance, even as training loss continues to fall.

Interestingly, the benefit of longer adaptation decreases with model size.
At 0.8B, the 100B checkpoint improves 5/7 benchmarks over its 50B counterpart.
At 9B, however, it improves only 1/7, while the 50B checkpoint performs better on the remaining 6/7.
The regressions are concentrated in knowledge and mathematics, whereas code more often benefits from continued adaptation, consistent with our code-heavy training mixture and the mixture sensitivity observed in~\cref{app:training}.
Thus, particularly at larger scales, useful AR-to-DLM adaptation emerges by 50B tokens, and further training under the same recipe can be counterproductive.

\subsection{Hybrid DLMs can decode in any order, even though RNNs are causal}
\begin{figure}[tp]
\centering
\fitwidth{\includegraphics{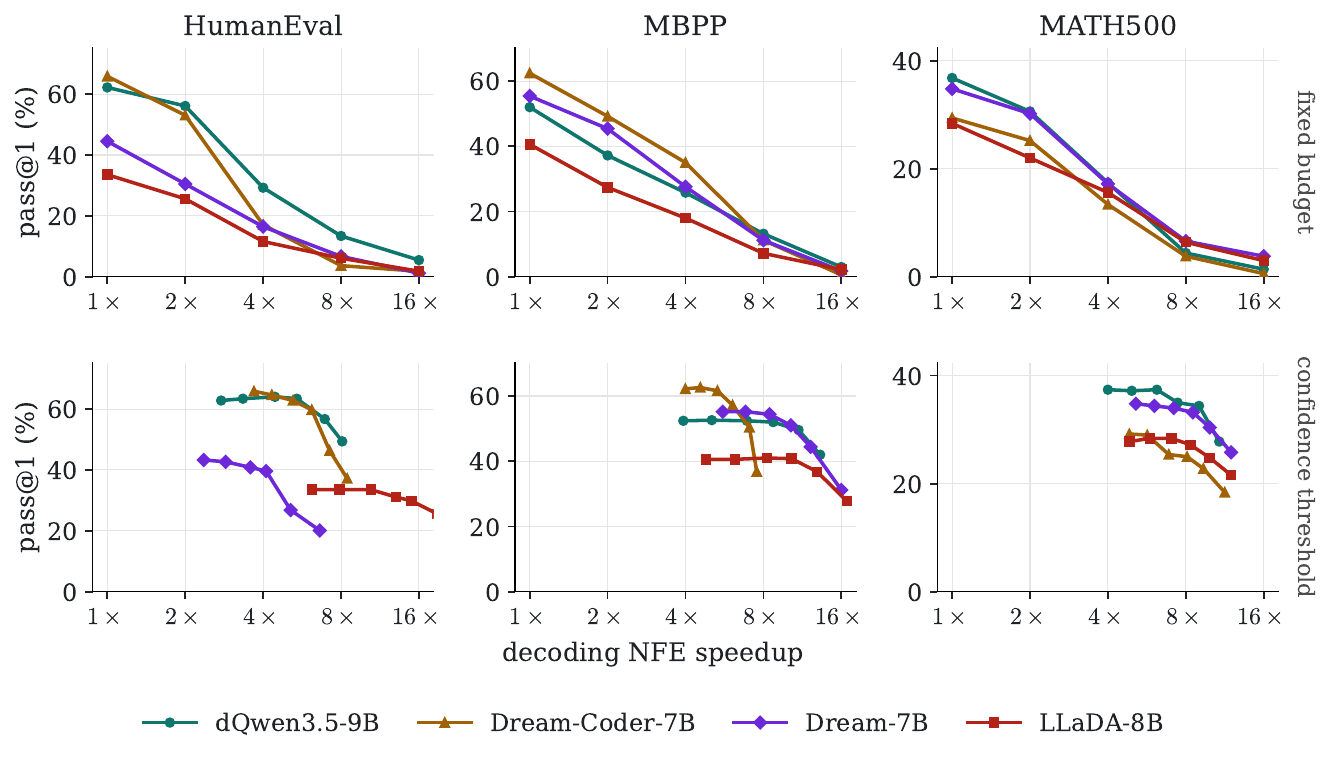}}

\caption{\textbf{Parallel decoding for the $6.5$--$7$B trunks.}
We use full-canvas decoding with size 1024; $1\times$ denotes one token per step, or 1024 NFEs, and larger speedups use proportionally fewer NFEs.
\textit{Top:} fixed decoding budgets, showing how performance degrades at faster speeds (fewer steps).
\textit{Bottom:} adaptive confidence thresholding with $\tau \in \{0.5, 0.6, 0.7, 0.8, 0.9, 0.95\}$.}
\label{fig:decode-7b}
\end{figure}

\label{sec:eval-order}

A central concern with retaining causal RNN layers is that the resulting DLM might simply inherit a left-to-right decode order.
We therefore study decode trajectories on HumanEval, recording the decoding step $d_k$ on which each canvas position $k$ is decoded.
We use a canvas of size 256 because HumanEval solutions are short, and a longer canvas would mostly measure padding (\cref{app:decodebehav}).
To distinguish local from global ordering, we define
\[
    \mathrm{ARL} = \frac{|\{k : d_k < d_{k+1}\}|}{|\{k : d_k \neq d_{k+1}\}|},
    \qquad
    \mathrm{ARG} = \frac{|\{i<j : d_i < d_j\}|}{|\{i<j : d_i \neq d_j\}|}.
\]
Here, local AR-ness ($\mathrm{ARL}$) measures left-to-right ordering among adjacent positions, while global AR-ness ($\mathrm{ARG}$) measures it across all position pairs.
Random order scores $0.5$ on both metrics, whereas strict AR decoding scores $1.0$.

As shown in~\cref{fig:decode-order}, all four DLMs are globally left-to-right but locally much more flexible.
Under full-canvas decoding, \dqfam{}-9B has local AR-ness $0.636$, squarely within the $0.631$--$0.652$ range of the three full-attention DLMs, despite retaining causal RNNs in most of its layers.
At the same time, all models have high global AR-ness, ranging from $0.884$ to $0.967$.
This local--global distinction is consistent with prior observations that even full-attention DLMs, including models trained from scratch, tend to globally decode from left to right~\citep{gong2026diffucoder}.
Thus, the causal RNN layers in \dqfam{} do not prevent the local out-of-order decoding behavior characteristic of full-attention DLMs.
We report additional decoding schemes, models, and trajectories in~\cref{app:decodebehav}.

\subsection{Hybrid DLMs can decode in parallel}
\begin{table}[t]
\centering

\caption{\textbf{\dqfam{} is competitive with DLMs trained on far more tokens.}
Size classes are set by trunk parameters.
We bold the best DLM number in each column of a class.
Other sizes in~\cref{tab:benchmark-grid-full}.}

\label{tab:benchmark-grid}
\footnotesize
\setlength{\tabcolsep}{5pt}
\begin{tabular}{lrlrrrrrrr}
\toprule
model & trunk & tokens & MMLU & GSM8K & MATH500 & HEval & HEval+ & MBPP & MBPP+ \\
\midrule
\multicolumn{10}{l}{\emph{${\approx}1.4$B trunk}} \\
Qwen3-1.7B & 1.41B & --- & 60.30 & 74.75 & 48.40 & 40.24 & 37.80 & 43.80 & 46.03 \\
Qwen3.5-2B & 1.37B & --- & 57.23 & 58.98 & 30.80 & 37.20 & 34.15 & 23.40 & 31.75 \\
CoDA & 1.41B & 200B & 26.21 & 2.65 & 3.80 & 25.00 & 20.12 & 33.80 & 41.01 \\
\dqctrl{}-1.7B & 1.41B & 50B & \textbf{51.30} & \textbf{42.23} & 13.60 & 34.76 & 33.54 & 30.40 & 38.89 \\
\dqctrl{}-1.7B & 1.41B & 100B & 48.12 & 39.35 & \textbf{15.80} & 41.46 & 37.20 & \textbf{35.40} & \textbf{46.83} \\
\dqfam{}-2B & 1.37B & 50B & 45.50 & 20.55 & 11.00 & 37.20 & 32.32 & 34.20 & 40.48 \\
\dqfam{}-2B & 1.37B & 100B & 41.55 & 19.79 & 9.20 & \textbf{43.90} & \textbf{38.41} & 33.20 & 43.12 \\
\midrule
\multicolumn{10}{l}{\emph{$6.5$--$7$B trunk}} \\
Qwen2.5-7B & 6.53B & --- & 74.13 & 79.61 & 40.80 & 55.49 & 48.78 & 64.40 & 69.84 \\
Qwen2.5-Coder-7B & 6.53B & --- & 68.07 & 78.77 & 37.20 & 59.76 & 53.66 & 68.40 & 71.43 \\
Qwen3.5-9B & 6.92B & --- & 69.85 & 87.34 & 48.80 & 68.90 & 62.80 & 55.00 & 63.76 \\
LLaDA-8B & 6.98B & 2.3T & 65.88 & 71.34 & 28.40 & 33.54 & 28.66 & 40.60 & 45.50 \\
Dream-7B & 6.53B & 580B & 71.45 & 74.91 & 34.80 & 44.51 & 38.41 & 55.40 & 57.67 \\
Dream-Coder-7B & 6.53B & 322B & 65.41 & 72.18 & 29.40 & \textbf{65.85} & 57.93 & \textbf{62.40} & \textbf{65.08} \\
\dqfam{}-9B & 6.92B & 50B & \textbf{74.57} & \textbf{74.98} & \textbf{37.40} & 64.02 & \textbf{59.15} & 50.20 & 58.20 \\
\dqfam{}-9B & 6.92B & 100B & 72.78 & 69.98 & 36.80 & 62.20 & 58.54 & 52.00 & 56.61 \\
\bottomrule
\end{tabular}
\end{table}

\label{sec:eval-decode}

Having established that causal RNN layers do not prevent local any-order decoding, we next ask whether this flexibility translates into parallel generation.
\Cref{fig:decode-7b} evaluates \dqfam{}-9B against the other $6.5$--$7$B-trunk DLMs as decoding is increasingly parallelized.
We consider both fixed budgets, which compare models at matched NFE counts, and confidence thresholding, which allows each model to choose how aggressively to decode in parallel.
Despite retaining a causal recurrent majority, \dqfam{}-9B maintains a strong accuracy--speed trade-off under both schemes.

Under fixed budgets, \dqfam{}-9B leads all comparators on HumanEval at every speedup beyond $1\times$, while remaining competitive on MBPP and MATH500.
Confidence thresholding yields the same broader picture, with \dqfam{}-9B remaining on a competitive quality--speed frontier across several-fold NFE speedups.
Thus, retaining causal recurrence in most of the backbone does not prevent the adapted model from realizing the parallel decoding behavior expected of a DLM.
Additional models and decoding settings are reported in~\cref{app:decode-figs}.





\subsection{\dqfam{} is a competitive model family}
\label{sec:eval-placement}

Having established that hybrid backbones adapt efficiently while attaining the decoding behavior expected of DLMs, we next ask how they compare on standard benchmarks.
\Cref{tab:benchmark-grid} evaluates base DLMs under the same full-canvas, one-token-per-step decoding scheme, grouped by trunk size.

Despite using substantially fewer training tokens, \dqfam{} is competitive across both comparison scales.
At the ${\approx}1.4$B trunk scale, \dqfam{}-2B after 50B adaptation tokens outperforms CoDA, trained for 200B tokens, on 6/7 benchmarks.
At the $6.5$--$7$B scale, \dqfam{}-9B after 50B tokens attains the best DLM score on 4/7 benchmarks against Dream-7B (580B tokens), Dream-Coder-7B (322B), and LLaDA-8B (2.3T).
Together with the preceding results, this suggests that retaining a mostly causal hybrid backbone need not trade away either DLM-like decoding behavior or downstream capability.
We report the remaining model sizes and block-decoding results in~\cref{app:block}.

\section{Discussion}
\label{sec:discussion}


\paragraph{Why might hybrid attention suffice for diffusion?}
Although each GDN is causal, bidirectional attention injects future context into representations processed by downstream layers.
This sparse bidirectional mixing may suffice if non-causal dependence in language is itself comparatively selective.
Suppose, for example, that language modeling consists mostly of causal dependencies, except for a small set of ``anchor'' tokens~\citep{rout2025anchored} that carry relevant future context, such as a later noun or verb that disambiguates an earlier phrase.
Let $A \subset [L]$ with $|A| \ll L$ denote such positions, and consider the factorization
$q(x_1, \ldots, x_L) = q(x_A)\prod_{i \notin A} q(x_i \mid x_{<i}, x_A)$.
Under this view, the $x_A$ tokens capture the non-causal dependence, while every other token is autoregressive in its prefix once $x_A$ is known.
The causal RNN layers can therefore model the prefix dependence, while the bidirectional attention layers need only provide access to a few future positions.
When $|A| \ll L$, this gives an interpretation for why a small number of bidirectional attention layers may suffice within a mostly causal recurrent backbone.


\paragraph{AR-to-DLM adaptation.}
AR-to-DLM adaptation remains expensive, with many models trained for hundreds of billions of additional tokens.
Our results suggest that the backbone itself can be one important factor: \dqfam{} changes the information flow in only a minority of its layers, while the recurrent majority remains causal.
In our matched-scale comparison, this hybrid reaches the same diffusion training loss in roughly half the tokens required by the full-attention control, suggesting that modern hybrid AR models may be particularly efficient starting points for adaptation.

\paragraph{Limitations.}
Our direct architectural comparison is limited to one pair at one scale, \dqfam{}-2B and \dqctrl{}-1.7B, whose trunks are closely matched but whose parent models and tokenizers differ.
Thus, the training-loss comparison is informative about adaptation speed, but downstream differences cannot be attributed solely to backbone architecture.
Our training mixture may also be further tuned, and our development experiments show that mixture choice can substantially alter what capabilities survive adaptation (\cref{app:training}).
Finally, we evaluate base DLMs only, as post-training (SFT, RL) substantially affects benchmark performance and is itself recipe-sensitive.



\paragraph{Future work.}
Our results suggest that useful adaptation may be possible at substantially less than 50B tokens, especially at larger scales (\cref{sec:eval-budget}).
A future study should pretrain matched hybrid and full-attention AR models from scratch, then adapt both under the same tokenizer, data, and optimization recipe.
This would isolate whether the faster adaptation we observe is genuinely caused by hybrid structure and clarify how the fraction and placement of RNN layers affect the trade-off.
More broadly, understanding which parts of an AR backbone are critical for diffusion language modeling will enable more efficient adaptations.

\paragraph{Acknowledgments.}
This research has been supported by NSF Grants 2505865 and 2112471, the UT Austin Machine Learning Lab, and computing support on the Vista GPU Cluster through the Center for Generative AI (CGAI) and the Texas Advanced Computing Center (TACC) at UT Austin.

\bibliography{references}
\bibliographystyle{plainnat}

\clearpage
\appendix
\section{Additional background on masked diffusion language models}
\label{app:objective}

\paragraph{Notation and forward process.}
Let $V$ be the vocabulary, and let $x=(x^1,\ldots,x^L)$ denote the clean tokens on an $L$-length generation canvas, with any prompt tokens held fixed and suppressed from the notation.
We represent each token $x^k$ and the mask token $m$ as one-hot vectors in $\mathbb{R}^{|V|}$.
Masked diffusion gradually replaces canvas tokens with $m$ according to a monotonically decreasing survival schedule $\alpha_t\in[0,1]$, with $\alpha_0=1$ and $\alpha_1=0$~\citep{austin2021structured,shi2024simplified,sahoo2024simple}.
The forward marginal factorizes across positions,
\[
q(z_t\mid x)
=
\prod_{k=1}^L q(z_t^k\mid x),
\qquad
q(z_t^k\mid x)
=
\mathrm{Cat}\!\left(
z_t^k;\,
\alpha_t x^k+(1-\alpha_t)m
\right).
\]
Thus, at time $t$, each canvas position remains clean with probability $\alpha_t$ and is masked with probability $1-\alpha_t$.
We use the linear schedule $\alpha_t=1-t$, so sampling $t\sim U(0,1)$ samples the masking ratio uniformly.

For a uniformly spaced $T$-step discretization of $[0,1]$, let
$t\in\{\tfrac1T,\tfrac2T,\ldots,1\}$ and $s=t-\tfrac1T$ denote the preceding time.
The corresponding forward transition is
\[
q(z_t^k\mid z_s^k)
=
\mathrm{Cat}\!\left(
z_t^k;\,
\frac{\alpha_t}{\alpha_s}z_s^k
+
\left(1-\frac{\alpha_t}{\alpha_s}\right)m
\right),
\qquad
z_0^k=x^k.
\]
Once a position becomes masked it remains masked, so $m$ is an absorbing state.

\paragraph{Reverse process.}
If the clean sequence $x$ were known, the exact reverse transition is
\[
q(z_s^k\mid z_t,x)
=
\begin{cases}
\mathrm{Cat}\!\left(z_s^k;\,z_t^k\right),
& z_t^k\neq m, \\[4pt]
\displaystyle
\mathrm{Cat}\!\left(
z_s^k;\,
\frac{\alpha_s-\alpha_t}{1-\alpha_t}x^k
+
\frac{1-\alpha_s}{1-\alpha_t}m
\right),
& z_t^k=m.
\end{cases}
\]
Thus, an unmasked position is copied unchanged, while a masked position is decoded to its clean value with probability
$(\alpha_s-\alpha_t)/(1-\alpha_t)$.

At generation time, $x$ is unknown, so the model predicts a clean-token distribution
$p_\theta^k(\cdot\mid z_t)\in\mathbb{R}^{|V|}$ at each position $k$.
Substituting this prediction for $x^k$ gives
\[
p_\theta(z_s\mid z_t)
:=
\prod_{k=1}^L p_\theta(z_s^k\mid z_t),
\qquad
p_\theta(z_s^k\mid z_t)
:=
\begin{cases}
\mathrm{Cat}\!\left(z_s^k;\,z_t^k\right),
& z_t^k\neq m, \\[4pt]
\displaystyle
\mathrm{Cat}\!\left(
z_s^k;\,
\frac{\alpha_s-\alpha_t}{1-\alpha_t}
p_\theta^k(\cdot\mid z_t)
+
\frac{1-\alpha_s}{1-\alpha_t}m
\right),
& z_t^k=m.
\end{cases}
\]
Generation begins with the prompt fixed and the $L$-length canvas fully masked, then repeatedly applies these reverse transitions until the canvas is decoded.

\paragraph{Training objective.}
For the absorbing-mask process above, the parameter-dependent part of the standard negative evidence lower bound (NELBO) reduces to
\begin{equation}
\label{eq:nelbo}
\mathcal{L}_{\mathrm{NELBO}}(x;\theta)
=
\sum_t
\mathbb{E}_{z_t\sim q(\cdot\mid x)}
\lambda_t
\sum_{k:z_t^k=m}
\left[-\log p_\theta^k(x^k\mid z_t)\right],
\qquad
\lambda_t
=
\frac{\alpha_s-\alpha_t}{1-\alpha_t},
\end{equation}
up to terms independent of $\theta$
\citep{austin2021structured,shi2024simplified,sahoo2024simple}.
Here, $\lambda_t$ is the probability that a masked position is decoded in the exact reverse transition from $t$ to $s$.

More generally, each time may be reweighted by a factor $w_t$,
\[
\mathcal{L}_{w}(x;\theta)
=
\sum_t
\mathbb{E}_{z_t\sim q(\cdot\mid x)}
w_t\lambda_t
\sum_{k:z_t^k=m}
\left[-\log p_\theta^k(x^k\mid z_t)\right].
\]
Following the \emph{simple weighting} of \citet{shi2025demystifying}, we take $w_t=1/\lambda_t$, so the schedule-dependent factor cancels and every masked target receives equal weight.
Under our linear schedule, we sample $t\sim U(0,1)$ directly and train with
\[
\mathcal{L}(x;\theta)
=
\mathbb{E}_{t\sim U(0,1)}
\mathbb{E}_{z_t\sim q(\cdot\mid x)}
\frac{1}{L}
\sum_{k:z_t^k=m}
\left[-\log p_\theta^k(x^k\mid z_t)\right].
\]
This is the training objective used throughout our adaptation experiments.
Because it differs from the standard NELBO, its value is not directly a likelihood bound \citep[App.~H.4]{shi2024simplified}.

\begin{remark*}
The standard NELBO places more per-target weight when the sequence is lightly masked.
Canceling $\lambda_t$ instead weights every masked target equally, shifting more total training weight toward highly masked states encountered early in generation.
Related reweightings retain a variational interpretation \citep{shi2025demystifying}.
\end{remark*}

\clearpage
\section{Training details}
\label{app:training}

\subsection{Training configuration}
\label{app:asrun}

\Cref{tab:mixture} gives the full training mixture with provenance, with subsets written as \texttt{parent\,/\,subset}.
Despite its name, \texttt{Nemotron-Pretraining-SFT-v1} contains SFT-\emph{style} data released for blending into pretraining, and we use it accordingly.
\Cref{tab:optim} gives the complete optimization configuration.

\begin{table}[!ht]
\centering
\caption{\textbf{Training data mixture.} Token shares of the 14 subsets, grouped by source repository within NVIDIA's Nemotron pretraining releases \citep{su2024nemotroncc,nvidia2025nemotronnano2,nvidia2025nemotron3nano,karimimahabadi2025nemotronccmath}; only the \texttt{nvidia/} organization prefix is elided.
The two \texttt{verified} rows are execution-verified derivations of \texttt{Code-Concepts}.}
\label{tab:mixture}
\small
\newcommand{\srcf}[1]{{\footnotesize\ttfamily #1}}
\begin{tabular}{llr}
\toprule
family & source & token share (\%) \\
\midrule
\multirow{10}{*}{code (50\%)}
 & \srcf{Nemotron-Pretraining-Code-v2\,/\,Synthetic-Code:} & \\
 & \qquad\srcf{synthetic-student-teacher} & 7 \\
 & \qquad\srcf{synthetic-rewriting} & 6 \\
 & \qquad\srcf{synthetic-question-answering} & 4 \\
 & \qquad\srcf{synthetic-code-review} & 3 \\
 & \srcf{Nemotron-Pretraining-Specialized-v1.1:} & \\
 & \qquad\srcf{Code-Concepts} {\footnotesize(verified, completion)} & 5 \\
 & \qquad\srcf{Code-Concepts} {\footnotesize(verified, instruct)} & 5 \\
 & \srcf{Nemotron-Pretraining-SFT-v1\,/\,SFT-Code} & 12 \\
 & \srcf{Nemotron-CC-Code-v1} & 8 \\
\midrule
\multirow{2}{*}{math (15\%)}
 & \srcf{Nemotron-Pretraining-Specialized-v1\,/\,Math-Textbooks} & 10 \\
 & \srcf{Nemotron-CC-Math-v1\,/\,4plus\_MIND} & 5 \\
\midrule
\multirow{6}{*}{general (35\%)}
 & \srcf{Nemotron-Pretraining-Specialized-v1:} & \\
 & \qquad\srcf{Wiki-Rewrite} & 5 \\
 & \qquad\srcf{InfiniByte-Reasoning} & 5 \\
 & \srcf{Nemotron-CC-v2.1:} & \\
 & \qquad\srcf{High-Quality} & 20 \\
 & \qquad\srcf{High-Quality-DQA} & 5 \\
\bottomrule
\end{tabular}
\end{table}

\begin{table}[t]
\centering
\caption{\textbf{Optimization configuration, as run.} Identical across sizes except the learning rate.
The 25k-step branch is the same schedule with decay starting at step 20{,}000.
Unlisted settings are HuggingFace Trainer defaults.}
\label{tab:optim}
\small
\begin{tabular}{ll}
\toprule
optimizer & AdamW ($\beta_1{=}0.9$, $\beta_2{=}0.95$) \\
weight decay & 0.01 \\
gradient clip & 1.0 \\
learning rate & 0.8B: $3{\times}10^{-5}$ \quad 2B: $2{\times}10^{-5}$ \quad
                4B: $2{\times}10^{-5}$ \quad 9B: $1{\times}10^{-5}$ \\
schedule & linear warmup (1k steps), stable (44k), cosine decay to 0 (5k) \\
batch & 512 sequences $\times$ 4096 tokens
        (${\sim}$2M content tokens/step) \\
precision & bf16 (tf32 matmul) \\
\bottomrule
\end{tabular}
\end{table}

\paragraph{Code preprocessing.}
For the Code-Concepts subset, we additionally verify the released solutions against examples provided in their docstrings.
We execute at most three declared examples per row and retain only solutions that pass, leaving 53.4\% of the original pool.
We render each surviving solution in two forms.
The \emph{completion form} maps a function signature to its body, matching tasks such as HumanEval.
Because MBPP instead provides a natural-language specification, we also construct an \emph{instruction form}: the docstring prose is spliced into a short instruction, with its verified examples appearing in a flush-left \texttt{>{>}>} preamble.

\paragraph{Augmentations.}
Sequences are packed to length 4096 and lightly augmented.
With probability 0.05, an entire batch is truncated to a shorter length; with probability 0.10, a sample retains a clean prefix; and with probability 0.05, we append a supervised padding suffix.
A quarter of the clean-prefix cases are instead trained fully masked, corresponding to ${\sim}2.5\%$ of rows overall.
Truncated lengths are sampled on a 256-token grid with mean 2048.
Together, these augmentations leave about 3.9k content tokens per row on average, or ${\sim}2$M content tokens in a 512-row step.
Thus, 25k and 50k training steps correspond to approximately 50B and 100B content tokens, respectively.

\subsection{Telemetry}
\label{app:telemetry}
\begin{figure}[tp]
\centering
\fitwidth{\includegraphics{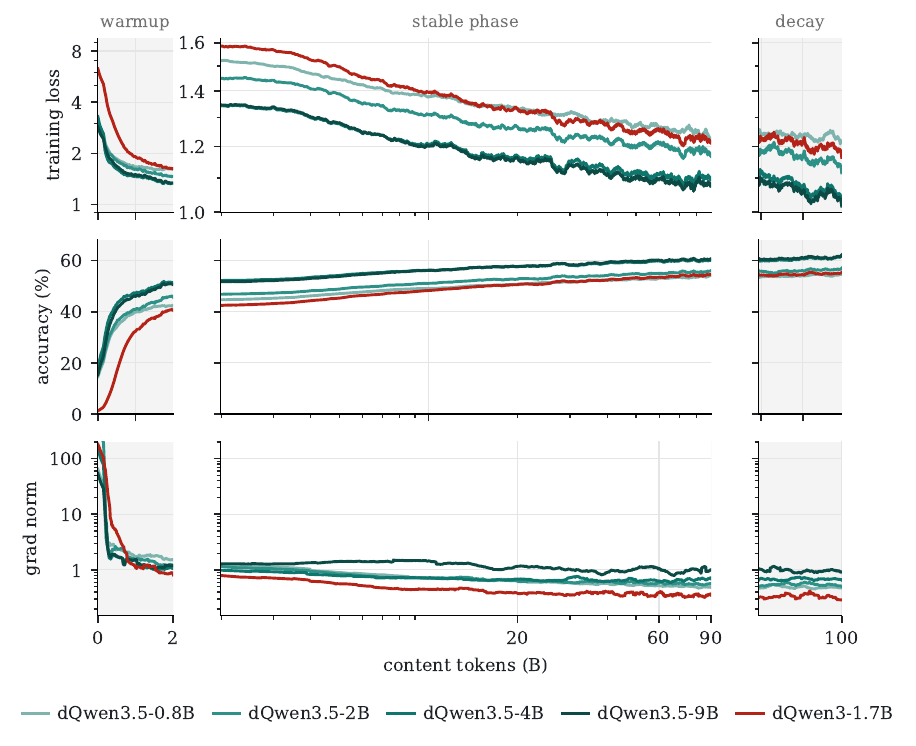}}
\caption{\textbf{Training telemetry for the released runs.} Loss, masked-token accuracy and pre-clip gradient norm for the four adapted sizes and the control, on aligned phase panels.
Accuracy uses our own masking distribution (\cref{app:asrun}), so its level is not comparable to other reports.}
\label{fig:telemetry}
\end{figure}

Every released run logs training loss, masked-token accuracy, and pre-clip gradient norm; \cref{fig:telemetry} shows all three over the full schedule.
The traces are smooth across runs, with no obvious optimization instability.
We include the gradient norm so that training stability can be inspected directly alongside the released telemetry.
Loss is still falling at the end of every run, since we train to fixed 50B- and 100B-token budgets rather than to convergence.
As discussed in~\cref{sec:recipe-dynamics}, longer adaptation can nevertheless degrade downstream performance.

\subsection{Development notes: learning rate and mixture}
\label{app:devnotes}

\paragraph{Learning-rate calibration.}
Adaptation is sensitive to learning rate in ways that training loss alone does not reveal.
In preliminary single-seed probes, we swept several learning rates and measured how much of each source model's above-chance MMLU performance survived adaptation; these probes use a different decoding setup and are not directly comparable with the main tables.
Notably, our initial $10^{-4}$ configuration caused the smaller models to lose nearly all of their above-chance MMLU performance, while the 9B model trained at a lower learning rate retained substantially more.
We interpret this pattern as catastrophic forgetting from overly aggressive updates, consistent with prior adaptation and continual-pretraining studies \citep{fu2025efficientdlm,ye2025dream,gupta2023continual}.
We therefore selected learning rates by capability retention rather than training loss.

\paragraph{Mixture exploration.}
Our initial mixture contained 71\% code and no general web text; we arrived at the released mixture through a series of small-scale probes.
At matched budgets, the released mixture improves 19/20 benchmark cells over this initial version, with code performance improving even as the code share falls to 50\%.
Thus, the larger initial code share was not translating into stronger code capability.
Mixture tuning was most impactful for smaller models.



\clearpage
\section{Extended evaluation}
\label{app:evalx}

\subsection{Evaluation details}
\label{app:evaldetails}

\Cref{tab:benchsetup} lists the per-benchmark configuration.
MMLU is multiple choice and is scored on the option's single answer token, while MATH500 uses the \texttt{lm-eval} harness's symbolic-equivalence check rather than string matching.
Unless otherwise specified, DLM generation uses a canvas of size 1024 with greedy highest-confidence-first unmasking and no remasking.
We use the same decoding scheme across models rather than each model's released inference recipe.

\begin{table}[!ht]
\centering
\caption{\textbf{Per-benchmark configuration.} Generation length is the canvas size for every generative task.
Problem counts are as graded.
The plus (\texttt{+}) variants feature harder unit tests; only HumanEval+ keeps its base benchmark's prompts.}
\label{tab:benchsetup}
\small
\begin{tabular}{llrrl}
\toprule
benchmark & task & shots & problems & scoring \\
\midrule
MMLU & \texttt{mmlu} & 5 & 14{,}042 & answer-token likelihood \\
GSM8K & \texttt{gsm8k\_cot} & 8 & 1{,}319 & exact match \\
MATH500 & \texttt{minerva\_math500} & 4 & 500 & symbolic equivalence \\
HumanEval & \texttt{humaneval} & 0 & 164 & pass@1 \\
HumanEval+ & \texttt{humaneval\_plus\_sound} & 0 & 164 & pass@1 \\
MBPP & \texttt{mbpp} & 3 & 500 & pass@1 \\
MBPP+ & \texttt{mbpp\_plus\_full} & 3 & 378 & pass@1 \\
MBPP (fence) & \texttt{mbpp\_ticks} & 3 & 500 & pass@1 \\
MBPP+ (fence) & \texttt{mbpp\_plus\_ticks} & 3 & 378 & pass@1 \\
\bottomrule
\end{tabular}
\end{table}

\FloatBarrier

\subsection{Checking our evaluation harness}
\label{app:reproduction}

\Cref{tab:reproduction} compares our scores with those reported by each comparator's authors for the same base checkpoints.
Across 18 referenced evaluations, the median absolute difference is about two points, supporting the use of our re-implemented evaluation harness.
Two code comparisons differ by roughly 13 points, Dream-7B on HumanEval and Dream-Coder-7B on MBPP, where their released evaluation scripts use different generation or post-processing choices.
We nevertheless use one fixed decoding scheme throughout the main comparison.

\begin{table}[!ht]
\centering
\caption{\textbf{A check on our evaluation harness.} We rescore each comparator under our own protocol and compare against what its authors published, to confirm our implementation is sound.
We fix one setting for everyone rather than each model's released sampler, which explains the gap.}
\label{tab:reproduction}
\small
\setlength{\tabcolsep}{3.4pt}
\begin{tabular}{l rrr rrr rrr rrr}
\toprule
 & \multicolumn{3}{c}{LLaDA-8B\rlap{$^{a}$}} & \multicolumn{3}{c}{Dream-7B\rlap{$^{b}$}}
 & \multicolumn{3}{c}{Dream-Coder-7B\rlap{$^{c}$}} & \multicolumn{3}{c}{CoDA\rlap{$^{d}$}} \\
\cmidrule(lr){2-4}\cmidrule(lr){5-7}\cmidrule(lr){8-10}\cmidrule(lr){11-13}
benchmark & pub. & ours & $\Delta$ & pub. & ours & $\Delta$ & pub. & ours & $\Delta$ & pub. & ours & $\Delta$ \\
\midrule
MMLU & 65.9 & 65.88 & $-0.0$ & 69.5 & 71.45 & $+2.0$ & 65.6 & 65.41 & $-0.2$ & --- & --- & --- \\
GSM8K & 70.3 & 71.34 & $+1.0$ & 77.2 & 74.91 & $-2.3$ & 71.1 & 72.18 & $+1.1$ & --- & --- & --- \\
HumanEval & 35.4 & 33.54 & $-1.9$ & 57.9 & 44.51 & $\mathbf{-13.4}$ & 66.5 & 65.85 & $-0.7$ & 29.3 & 25.00 & $-4.3$ \\
HumanEval+ & --- & --- & --- & --- & --- & --- & 60.4 & 57.93 & $-2.5$ & 23.8 & 20.12 & $-3.7$ \\
MBPP & 40.0 & 40.60 & $+0.6$ & 56.2 & 55.40 & $-0.8$ & 75.9 & 62.40 & $\mathbf{-13.5}$ & 35.2 & 33.80 & $-1.4$ \\
MBPP+ & --- & --- & --- & --- & --- & --- & 61.6 & 65.08 & $+3.5$ & 46.0 & 41.01 & $-5.0$ \\
\bottomrule
\end{tabular}

\vspace{3pt}
{\footnotesize
\begin{minipage}{\linewidth}
Ours: MMLU 5-shot, GSM8K 8, HumanEval 0, MBPP 3.
An empty cell means the paper does not report it.

\vspace{1.5pt}
$^{a}$~\texttt{arXiv:2502.09992v3}, Table 1.
Generates on a 1024-position canvas at one token per step, the closest of the four to our protocol; GSM8K and MBPP at 4 shots.

$^{b}$~\texttt{arXiv:2508.15487v1}, Table 1.
Generates 512 tokens at a sampling temperature, with stop-string truncation disabled and an external sanitizer on HumanEval; MBPP at 4 shots.

$^{c}$~\texttt{arXiv:2509.01142v1}, Table 1.
Generates 512 tokens zero-shot, and scores MBPP on the 378-problem EvalPlus set rather than the 500-problem split we use.

$^{d}$~\texttt{arXiv:2510.03270v1}, Table 4.
Caps generation at 768 tokens, zero-shot.
\end{minipage}}
\end{table}

\FloatBarrier

\subsection{Comparison with AR parents}
\label{app:ar-fractions}

\Cref{tab:ar-fractions} reports each adapted DLM as a fraction of its own AR parent, helping separate capability inherited from the parent model from capability retained through adaptation.
For example, \dqfam{}-9B's lower GSM8K performance is not inherited from a weaker parent, since Qwen3.5-9B is the strongest AR parent in that comparison.
On MBPP, the picture reverses: Qwen3.5-9B begins behind the other AR parents, while \dqfam{}-9B retains more of its parent's performance.

\begin{table}[!ht]
\centering
\caption{\textbf{Each adapted DLM relative to its AR parent.}
We divide each DLM score by the corresponding AR parent's score; for our models, we use the 100B checkpoints.
Values above 1 indicate higher benchmark performance after adaptation.}

\label{tab:ar-fractions}
\footnotesize
\begin{tabular}{llrrrrrrr}
\toprule
model & AR parent & MMLU & GSM8K & MATH500 & HEval & HEval+ & MBPP & MBPP+ \\
\midrule
\multicolumn{9}{l}{\emph{${\approx}0.5$B trunk}} \\
\dqfam{}-0.8B & Qwen3.5-0.8B & 0.601 & 0.232 & 0.365 & 1.568 & 1.530 & 1.333 & 1.341 \\
\midrule
\multicolumn{9}{l}{\emph{${\approx}1.4$B trunk}} \\
CoDA & Qwen3-1.7B & 0.435 & 0.035 & 0.079 & 0.621 & 0.532 & 0.772 & 0.891 \\
\dqctrl{}-1.7B & Qwen3-1.7B & 0.798 & 0.526 & 0.326 & 1.030 & 0.984 & 0.808 & 1.017 \\
\dqfam{}-2B & Qwen3.5-2B & 0.726 & 0.336 & 0.299 & 1.180 & 1.125 & 1.419 & 1.358 \\
\midrule
\multicolumn{9}{l}{\emph{${\approx}3.6$B trunk}} \\
\dqfam{}-4B & Qwen3.5-4B & 0.830 & 0.627 & 0.516 & 1.021 & 1.068 & 1.109 & 1.033 \\
\midrule
\multicolumn{9}{l}{\emph{$6.5$--$7$B trunk}} \\
Dream-7B & Qwen2.5-7B & 0.964 & 0.941 & 0.853 & 0.802 & 0.787 & 0.860 & 0.826 \\
Dream-Coder-7B & Qwen2.5-Coder-7B & 0.961 & 0.916 & 0.790 & 1.102 & 1.080 & 0.912 & 0.911 \\
\dqfam{}-9B & Qwen3.5-9B & 1.042 & 0.801 & 0.754 & 0.903 & 0.932 & 0.945 & 0.888 \\
\bottomrule
\end{tabular}
\end{table}

\FloatBarrier

\subsection{Prompt-format sensitivity}
\label{app:protosens}

MBPP's \texttt{[BEGIN]}/\texttt{[DONE]} scaffold is specific to its evaluation harness, so we rescore every DLM using a plain \texttt{\textasciigrave\textasciigrave\textasciigrave python} fence while holding the remaining decoding scheme fixed (\cref{tab:format-sensitivity}).
Prompt format changes scores modestly overall, with larger effects for smaller models.
The relative picture is broadly unchanged, and the main text reports the default scaffold throughout.

\begin{table}[!ht]
\centering
\caption{\textbf{MBPP is sensitive to its prompt format, and smaller models more so.} Each model scored under the stock \texttt{[BEGIN]}/\texttt{[DONE]} scaffold and under a plain \texttt{\textasciigrave\textasciigrave\textasciigrave python} fence, everything else held fixed.
The gain from the fence falls with model size and reverses for the largest models.}
\label{tab:format-sensitivity}
\small
\begin{tabular}{lrlrrrrrr}
\toprule
& & & \multicolumn{3}{c}{MBPP} & \multicolumn{3}{c}{MBPP+} \\
\cmidrule(lr){4-6}\cmidrule(lr){7-9}
model & trunk & tokens & stock & fence & $\Delta$ & stock & fence & $\Delta$ \\
\midrule
\multicolumn{9}{l}{\emph{${\approx}0.5$B trunk}} \\
\dqfam{}-0.8B & 0.50B & 50B & 22.60 & 23.80 & +1.20 & 29.10 & 33.33 & +4.23 \\
\dqfam{}-0.8B & 0.50B & 100B & 23.20 & 29.40 & +6.20 & 32.28 & 37.04 & +4.76 \\
\midrule
\multicolumn{9}{l}{\emph{${\approx}1.4$B trunk}} \\
CoDA & 1.41B & 200B & 33.80 & 37.00 & +3.20 & 41.01 & 43.65 & +2.64 \\
\dqctrl{}-1.7B & 1.41B & 50B & 30.40 & 35.40 & +5.00 & 38.89 & 42.33 & +3.44 \\
\dqctrl{}-1.7B & 1.41B & 100B & 35.40 & 36.60 & +1.20 & 46.83 & 49.47 & +2.64 \\
\dqfam{}-2B & 1.37B & 50B & 34.20 & 34.40 & +0.20 & 40.48 & 42.59 & +2.11 \\
\dqfam{}-2B & 1.37B & 100B & 33.20 & 36.00 & +2.80 & 43.12 & 44.71 & +1.59 \\
\midrule
\multicolumn{9}{l}{\emph{${\approx}3.6$B trunk}} \\
\dqfam{}-4B & 3.57B & 50B & 44.60 & 46.80 & +2.20 & 51.32 & 53.70 & +2.38 \\
\dqfam{}-4B & 3.57B & 100B & 44.60 & 46.80 & +2.20 & 50.00 & 52.38 & +2.38 \\
\midrule
\multicolumn{9}{l}{\emph{$6.5$--$7$B trunk}} \\
LLaDA-8B & 6.98B & 2.3T & 40.60 & 39.20 & -1.40 & 45.50 & 44.71 & -0.79 \\
Dream-7B & 6.53B & 580B & 55.40 & 55.80 & +0.40 & 57.67 & 59.52 & +1.85 \\
Dream-Coder-7B & 6.53B & 322B & 62.40 & 63.80 & +1.40 & 65.08 & 64.81 & -0.27 \\
\dqfam{}-9B & 6.92B & 50B & 50.20 & 52.20 & +2.00 & 58.20 & 58.20 & +0.00 \\
\dqfam{}-9B & 6.92B & 100B & 52.00 & 51.00 & -1.00 & 56.61 & 59.79 & +3.18 \\
\bottomrule
\end{tabular}
\end{table}

\FloatBarrier

\clearpage
\subsection{Full-canvas and block decoding, one token per step}
\label{app:block}

\Cref{tab:benchmark-grid-full} gives the complete full-canvas results behind \cref{tab:benchmark-grid}, while \cref{tab:benchmark-grid-block} repeats the evaluation using 32-token blocks decoded in 32 steps each.
The two decoding schemes give very similar accuracy and preserve the model ordering, suggesting that our main comparisons are not particular to full-canvas decoding.
\Citet{nie2025large} similarly find that block diffusion sampling benefits their instruction-tuned model but not the base model, which is the setting considered here.

\begin{table}[!ht]
\centering
\caption{\textbf{Evaluations for full-canvas decoding at one token per step.}
We bold the best DLM number in each column of a size class.
Classes with a single size family are left unmarked.
}
\label{tab:benchmark-grid-full}
\footnotesize
\setlength{\tabcolsep}{5pt}
\begin{tabular}{lrlrrrrrrr}
\toprule
model & trunk & tokens & MMLU & GSM8K & MATH500 & HEval & HEval+ & MBPP & MBPP+ \\
\midrule
\multicolumn{10}{l}{\emph{${\approx}0.5$B trunk}} \\
Qwen3.5-0.8B & 0.50B & --- & 50.31 & 34.95 & 17.00 & 22.56 & 20.73 & 17.40 & 24.07 \\
\dqfam{}-0.8B & 0.50B & 50B & 32.62 & 8.49 & 5.40 & 28.05 & 25.61 & 22.60 & 29.10 \\
\dqfam{}-0.8B & 0.50B & 100B & 30.24 & 8.11 & 6.20 & 35.37 & 31.71 & 23.20 & 32.28 \\
\midrule
\multicolumn{10}{l}{\emph{${\approx}1.4$B trunk}} \\
Qwen3-1.7B & 1.41B & --- & 60.30 & 74.75 & 48.40 & 40.24 & 37.80 & 43.80 & 46.03 \\
Qwen3.5-2B & 1.37B & --- & 57.23 & 58.98 & 30.80 & 37.20 & 34.15 & 23.40 & 31.75 \\
CoDA & 1.41B & 200B & 26.21 & 2.65 & 3.80 & 25.00 & 20.12 & 33.80 & 41.01 \\
\dqctrl{}-1.7B & 1.41B & 50B & \textbf{51.30} & \textbf{42.23} & 13.60 & 34.76 & 33.54 & 30.40 & 38.89 \\
\dqctrl{}-1.7B & 1.41B & 100B & 48.12 & 39.35 & \textbf{15.80} & 41.46 & 37.20 & \textbf{35.40} & \textbf{46.83} \\
\dqfam{}-2B & 1.37B & 50B & 45.50 & 20.55 & 11.00 & 37.20 & 32.32 & 34.20 & 40.48 \\
\dqfam{}-2B & 1.37B & 100B & 41.55 & 19.79 & 9.20 & \textbf{43.90} & \textbf{38.41} & 33.20 & 43.12 \\
\midrule
\multicolumn{10}{l}{\emph{${\approx}3.6$B trunk}} \\
Qwen3.5-4B & 3.57B & --- & 69.82 & 85.29 & 49.60 & 59.15 & 53.66 & 40.20 & 48.41 \\
\dqfam{}-4B & 3.57B & 50B & 62.42 & 57.77 & 25.00 & 54.88 & 51.83 & 44.60 & 51.32 \\
\dqfam{}-4B & 3.57B & 100B & 57.95 & 53.45 & 25.60 & 60.37 & 57.32 & 44.60 & 50.00 \\
\midrule
\multicolumn{10}{l}{\emph{$6.5$--$7$B trunk}} \\
Qwen2.5-7B & 6.53B & --- & 74.13 & 79.61 & 40.80 & 55.49 & 48.78 & 64.40 & 69.84 \\
Qwen2.5-Coder-7B & 6.53B & --- & 68.07 & 78.77 & 37.20 & 59.76 & 53.66 & 68.40 & 71.43 \\
Qwen3.5-9B & 6.92B & --- & 69.85 & 87.34 & 48.80 & 68.90 & 62.80 & 55.00 & 63.76 \\
LLaDA-8B & 6.98B & 2.3T & 65.88 & 71.34 & 28.40 & 33.54 & 28.66 & 40.60 & 45.50 \\
Dream-7B & 6.53B & 580B & 71.45 & 74.91 & 34.80 & 44.51 & 38.41 & 55.40 & 57.67 \\
Dream-Coder-7B & 6.53B & 322B & 65.41 & 72.18 & 29.40 & \textbf{65.85} & 57.93 & \textbf{62.40} & \textbf{65.08} \\
\dqfam{}-9B & 6.92B & 50B & \textbf{74.57} & \textbf{74.98} & \textbf{37.40} & 64.02 & \textbf{59.15} & 50.20 & 58.20 \\
\dqfam{}-9B & 6.92B & 100B & 72.78 & 69.98 & 36.80 & 62.20 & 58.54 & 52.00 & 56.61 \\
\bottomrule
\end{tabular}
\end{table}

\begin{table}[!ht]
\centering
\caption{\textbf{Evaluations for block decoding at one token per step.}
We bold the best DLM number in each column of a size class.
Classes with a single size family are left unmarked.
}
\label{tab:benchmark-grid-block}
\footnotesize
\setlength{\tabcolsep}{5pt}
\begin{tabular}{lrlrrrrrrr}
\toprule
model & trunk & tokens & MMLU & GSM8K & MATH500 & HEval & HEval+ & MBPP & MBPP+ \\
\midrule
\multicolumn{10}{l}{\emph{${\approx}0.5$B trunk}} \\
Qwen3.5-0.8B & 0.50B & --- & 50.31 & 34.95 & 17.00 & 22.56 & 20.73 & 17.40 & 24.07 \\
\dqfam{}-0.8B & 0.50B & 50B & 32.62 & 8.49 & 5.20 & 28.05 & 25.61 & 22.20 & 29.63 \\
\dqfam{}-0.8B & 0.50B & 100B & 30.24 & 7.88 & 7.40 & 34.15 & 30.49 & 22.60 & 32.28 \\
\midrule
\multicolumn{10}{l}{\emph{${\approx}1.4$B trunk}} \\
Qwen3-1.7B & 1.41B & --- & 60.30 & 74.75 & 48.40 & 40.24 & 37.80 & 43.80 & 46.03 \\
Qwen3.5-2B & 1.37B & --- & 57.23 & 58.98 & 30.80 & 37.20 & 34.15 & 23.40 & 31.75 \\
CoDA & 1.41B & 200B & 26.21 & 3.18 & 3.20 & 23.17 & 17.07 & 35.00 & 42.33 \\
\dqctrl{}-1.7B & 1.41B & 50B & \textbf{51.30} & \textbf{41.55} & 14.40 & 35.98 & 33.54 & 31.20 & 40.21 \\
\dqctrl{}-1.7B & 1.41B & 100B & 48.12 & 39.95 & \textbf{15.80} & \textbf{43.29} & \textbf{37.80} & \textbf{36.60} & \textbf{47.35} \\
\dqfam{}-2B & 1.37B & 50B & 45.50 & 20.92 & 11.20 & 39.02 & 34.15 & 34.00 & 40.48 \\
\dqfam{}-2B & 1.37B & 100B & 41.55 & 19.41 & 10.40 & \textbf{43.29} & \textbf{37.80} & 34.40 & 44.44 \\
\midrule
\multicolumn{10}{l}{\emph{${\approx}3.6$B trunk}} \\
Qwen3.5-4B & 3.57B & --- & 69.82 & 85.29 & 49.60 & 59.15 & 53.66 & 40.20 & 48.41 \\
\dqfam{}-4B & 3.57B & 50B & 62.42 & 57.77 & 26.60 & 56.10 & 52.44 & 46.20 & 52.65 \\
\dqfam{}-4B & 3.57B & 100B & 57.95 & 54.66 & 25.80 & 59.15 & 55.49 & 48.00 & 51.06 \\
\midrule
\multicolumn{10}{l}{\emph{$6.5$--$7$B trunk}} \\
Qwen2.5-7B & 6.53B & --- & 74.13 & 79.61 & 40.80 & 55.49 & 48.78 & 64.40 & 69.84 \\
Qwen2.5-Coder-7B & 6.53B & --- & 68.07 & 78.77 & 37.20 & 59.76 & 53.66 & 68.40 & 71.43 \\
Qwen3.5-9B & 6.92B & --- & 69.85 & 87.34 & 48.80 & 68.90 & 62.80 & 55.00 & 63.76 \\
LLaDA-8B & 6.98B & 2.3T & 65.88 & 72.48 & 27.20 & 32.93 & 28.05 & 40.40 & 45.24 \\
Dream-7B & 6.53B & 580B & 71.45 & \textbf{75.89} & 35.00 & 43.90 & 39.02 & 56.60 & 58.73 \\
Dream-Coder-7B & 6.53B & 322B & 65.41 & 72.18 & 30.20 & \textbf{67.07} & 58.54 & \textbf{62.00} & \textbf{64.81} \\
\dqfam{}-9B & 6.92B & 50B & \textbf{74.57} & 74.91 & \textbf{38.20} & 62.20 & 57.32 & 51.60 & 58.47 \\
\dqfam{}-9B & 6.92B & 100B & 72.78 & 70.58 & 37.40 & 62.80 & \textbf{59.15} & 51.00 & 58.47 \\
\bottomrule
\end{tabular}
\end{table}

\FloatBarrier

\clearpage
\subsection{Additional decode order experiments}
\label{app:decodebehav}

\begin{figure}[p]
\centering
\fitwidth{\includegraphics{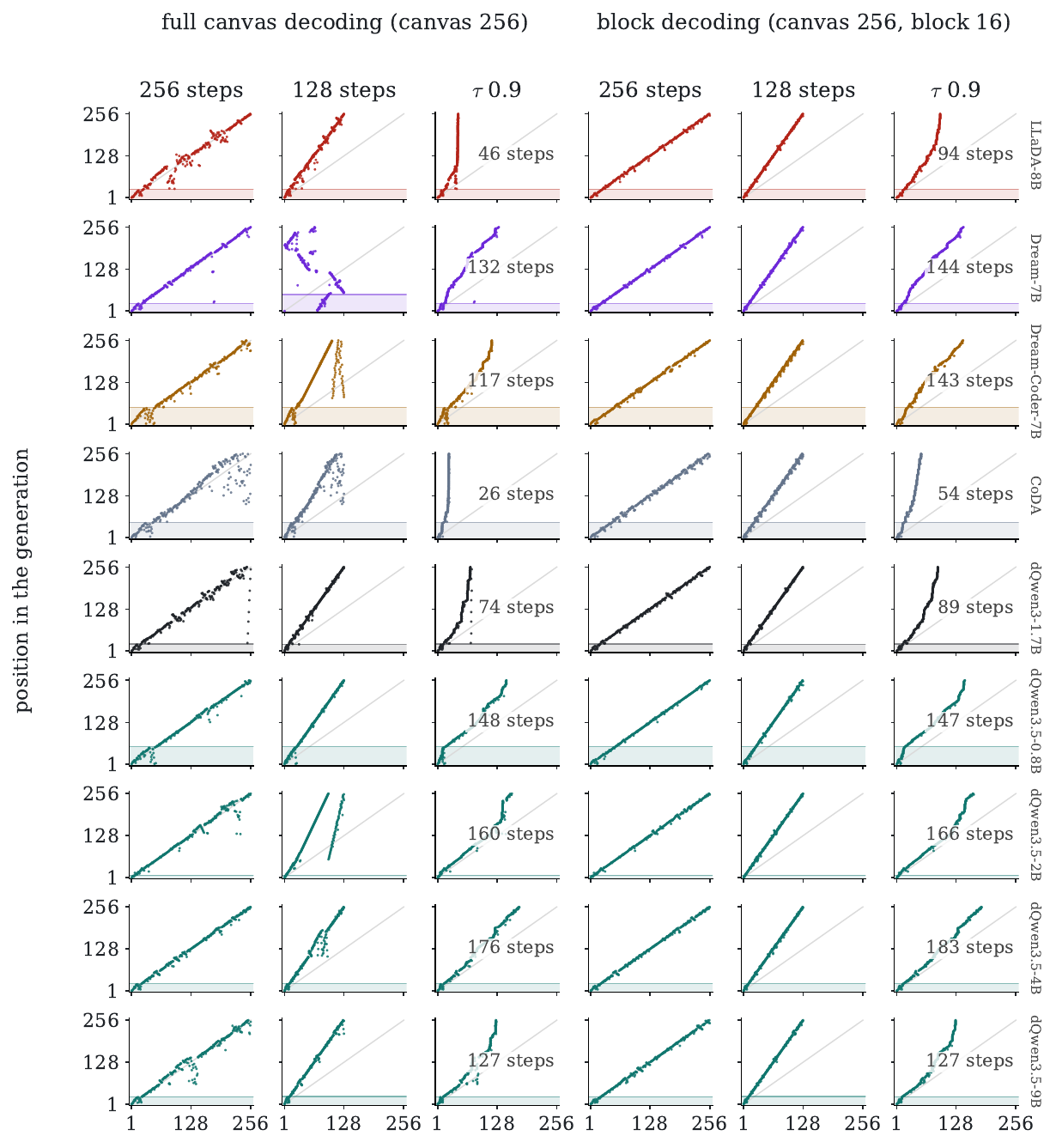}}
\caption{\textbf{Decode order on \texttt{HumanEval/27}, all evaluated DLMs.}
Each dot marks when and where a position was unmasked.}
\label{fig:decode-he27}
\end{figure}

We extend the HumanEval decode-order analysis of~\cref{sec:eval-order} across all evaluated DLMs and both full-canvas and block decoding.
We use canvas size 256 throughout and block size 16 for block decoding.
\Cref{tab:decode-order-full} shows that block decoding drives global AR-ness close to $1$ by construction, while local AR-ness remains substantially lower, preserving the same local--global distinction observed under full-canvas decoding.
The values cited in~\cref{sec:intro} use the full-canvas 256-step setting, and additional trajectories are shown in~\cref{fig:decode-he27}.

\begin{table}[!ht]
\centering
\caption{\textbf{Decode order for all evaluated DLMs, over 164 HumanEval problems.}
Local ($\mathrm{ARL}$) and global ($\mathrm{ARG}$) AR-ness under each scheme, with NFEs where a threshold sets them.
An autoregressive decoder scores 1.000 on both and a random order 0.500.}
\label{tab:decode-order-full}
\scriptsize
\setlength{\tabcolsep}{3.2pt}
\begin{tabular}{lccccccccccccccc}
\toprule
 & \multicolumn{2}{c}{full, 256 steps} & \multicolumn{2}{c}{full, 128 steps}
 & \multicolumn{3}{c}{full, $\tau = 0.9$} & \multicolumn{2}{c}{block, 256 steps}
 & \multicolumn{2}{c}{block, 128 steps} & \multicolumn{3}{c}{block, $\tau = 0.9$} \\
\cmidrule(lr){2-3}\cmidrule(lr){4-5}\cmidrule(lr){6-8}
\cmidrule(lr){9-10}\cmidrule(lr){11-12}\cmidrule(lr){13-15}
model & $\mathrm{ARL}$ & $\mathrm{ARG}$ & $\mathrm{ARL}$ & $\mathrm{ARG}$ & $\mathrm{ARL}$ & $\mathrm{ARG}$ & NFE & $\mathrm{ARL}$ & $\mathrm{ARG}$ & $\mathrm{ARL}$ & $\mathrm{ARG}$ & $\mathrm{ARL}$ & $\mathrm{ARG}$ & NFE \\
\midrule
LLaDA-8B & 0.631 & 0.932 & 0.617 & 0.924 & 0.630 & 0.945 & $83 \pm 38$ & 0.679 & 0.990 & 0.668 & 0.991 & 0.718 & 0.995 & $97 \pm 39$ \\
Dream-7B & 0.638 & 0.884 & 0.619 & 0.830 & 0.659 & 0.891 & $139 \pm 40$ & 0.706 & 0.992 & 0.708 & 0.993 & 0.739 & 0.995 & $147 \pm 39$ \\
Dream-Coder-7B & 0.652 & 0.937 & 0.607 & 0.934 & 0.662 & 0.944 & $136 \pm 55$ & 0.689 & 0.991 & 0.680 & 0.992 & 0.713 & 0.994 & $139 \pm 47$ \\
CoDA & 0.583 & 0.928 & 0.593 & 0.933 & 0.609 & 0.952 & $84 \pm 47$ & 0.625 & 0.987 & 0.643 & 0.989 & 0.701 & 0.995 & $94 \pm 41$ \\
\dqctrl{}-1.7B & 0.657 & 0.974 & 0.652 & 0.966 & 0.679 & 0.980 & $145 \pm 35$ & 0.695 & 0.992 & 0.699 & 0.993 & 0.723 & 0.995 & $149 \pm 33$ \\
\dqfam{}-0.8B & 0.642 & 0.974 & 0.651 & 0.971 & 0.651 & 0.979 & $144 \pm 35$ & 0.681 & 0.991 & 0.693 & 0.993 & 0.700 & 0.994 & $147 \pm 32$ \\
\dqfam{}-2B & 0.646 & 0.971 & 0.656 & 0.970 & 0.656 & 0.977 & $140 \pm 41$ & 0.686 & 0.991 & 0.703 & 0.993 & 0.710 & 0.995 & $146 \pm 37$ \\
\dqfam{}-4B & 0.641 & 0.972 & 0.644 & 0.969 & 0.657 & 0.978 & $130 \pm 36$ & 0.688 & 0.991 & 0.689 & 0.993 & 0.716 & 0.995 & $135 \pm 34$ \\
\dqfam{}-9B & 0.636 & 0.967 & 0.651 & 0.970 & 0.655 & 0.975 & $129 \pm 34$ & 0.691 & 0.991 & 0.703 & 0.993 & 0.721 & 0.995 & $133 \pm 34$ \\
\midrule
AR decoder & 1.000 & 1.000 & 1.000 & 1.000 & 1.000 & 1.000 & $256$ & 1.000 & 1.000 & 1.000 & 1.000 & 1.000 & 1.000 & $256$ \\
\bottomrule
\end{tabular}
\normalsize

\end{table}

\FloatBarrier

\FloatBarrier

\clearpage
\subsection{Additional parallelization experiments}
\label{app:decode-figs}

\begin{figure}[!ht]
\centering
\fitwidth{\includegraphics{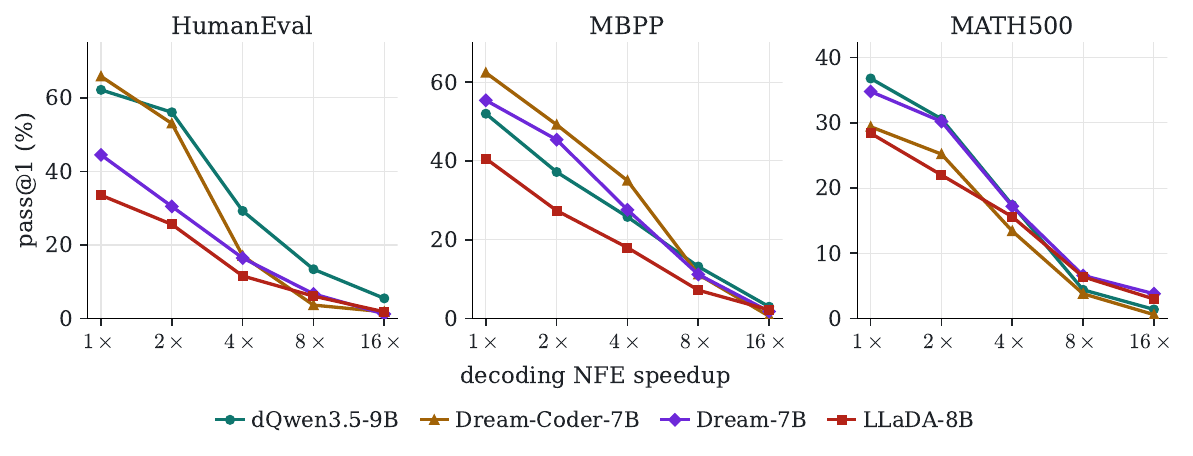}}
\\[6pt] \scriptsize
\setlength{\tabcolsep}{3pt}
\begin{tabular}{lrrrrrrrrrrrrrrr}
\toprule
 & \multicolumn{5}{c}{HumanEval} & \multicolumn{5}{c}{MBPP} & \multicolumn{5}{c}{MATH500} \\
\cmidrule(lr){2-6}\cmidrule(lr){7-11}\cmidrule(lr){12-16}
model & $1\times$ & $2\times$ & $4\times$ & $8\times$ & $16\times$ & $1\times$ & $2\times$ & $4\times$ & $8\times$ & $16\times$ & $1\times$ & $2\times$ & $4\times$ & $8\times$ & $16\times$ \\
\midrule
dQwen3.5-9B & 62.20 & 56.10 & 29.27 & 13.41 & 5.49 & 52.00 & 37.20 & 25.80 & 13.20 & 3.00 & 36.80 & 30.60 & 17.40 & 4.40 & 1.40 \\
Dream-Coder-7B & 65.85 & 53.05 & 17.07 & 3.66 & 1.83 & 62.40 & 49.20 & 35.00 & 11.20 & 0.60 & 29.40 & 25.20 & 13.40 & 3.80 & 0.60 \\
Dream-7B & 44.51 & 30.49 & 16.46 & 6.71 & 1.22 & 55.40 & 45.40 & 27.60 & 11.20 & 1.80 & 34.80 & 30.20 & 17.20 & 6.60 & 3.80 \\
LLaDA-8B & 33.54 & 25.61 & 11.59 & 6.10 & 1.83 & 40.60 & 27.40 & 18.00 & 7.20 & 2.20 & 28.40 & 22.00 & 15.60 & 6.40 & 3.00 \\
\bottomrule
\end{tabular}
\normalsize
\caption{\textbf{Parallel decoding at fixed step budgets, $6.5$--$7$B trunks.}}
\label{fig:app-decode-7b-fixed}
\end{figure}

\begin{figure}[!tp]
\centering
\fitwidth{\includegraphics{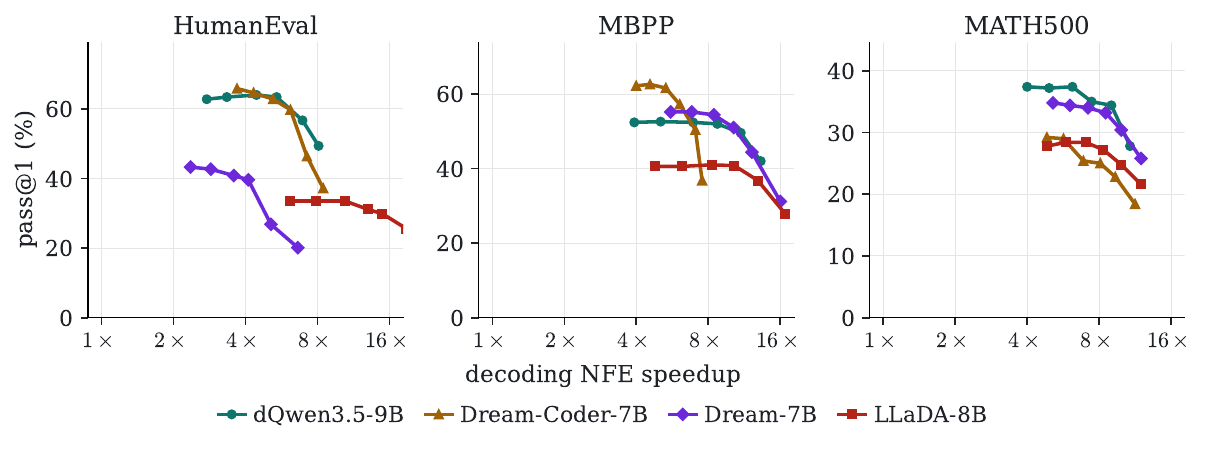}}
\\[6pt] \scriptsize
\setlength{\tabcolsep}{2pt}
\begin{tabular}{lrrrrrrrrrrrrrrrrrr}
\toprule
 & \multicolumn{6}{c}{HumanEval} & \multicolumn{6}{c}{MBPP} & \multicolumn{6}{c}{MATH500} \\
\cmidrule(lr){2-7}\cmidrule(lr){8-13}\cmidrule(lr){14-19}
model & 0.95 & 0.9 & 0.8 & 0.7 & 0.6 & 0.5 & 0.95 & 0.9 & 0.8 & 0.7 & 0.6 & 0.5 & 0.95 & 0.9 & 0.8 & 0.7 & 0.6 & 0.5 \\
\midrule
dQwen3.5-9B & \stackcell{62.80}{$2.8\times$} & \stackcell{63.41}{$3.3\times$} & \stackcell{64.02}{$4.4\times$} & \stackcell{63.41}{$5.4\times$} & \stackcell{56.71}{$6.9\times$} & \stackcell{49.39}{$8.1\times$} & \stackcell{52.40}{$3.9\times$} & \stackcell{52.60}{$5.1\times$} & \stackcell{52.40}{$6.9\times$} & \stackcell{52.00}{$8.7\times$} & \stackcell{49.60}{$10.9\times$} & \stackcell{42.00}{$13.3\times$} & \stackcell{37.40}{$4.0\times$} & \stackcell{37.20}{$4.9\times$} & \stackcell{37.40}{$6.2\times$} & \stackcell{35.00}{$7.4\times$} & \stackcell{34.40}{$9.0\times$} & \stackcell{27.80}{$10.8\times$} \\
Dream-Coder-7B & \stackcell{65.85}{$3.7\times$} & \stackcell{64.63}{$4.3\times$} & \stackcell{62.80}{$5.2\times$} & \stackcell{59.76}{$6.2\times$} & \stackcell{46.34}{$7.2\times$} & \stackcell{37.20}{$8.4\times$} & \stackcell{62.20}{$4.0\times$} & \stackcell{62.60}{$4.6\times$} & \stackcell{61.60}{$5.3\times$} & \stackcell{57.20}{$6.1\times$} & \stackcell{50.40}{$7.1\times$} & \stackcell{36.80}{$7.5\times$} & \stackcell{29.20}{$4.8\times$} & \stackcell{29.00}{$5.7\times$} & \stackcell{25.40}{$6.9\times$} & \stackcell{25.00}{$8.1\times$} & \stackcell{22.80}{$9.4\times$} & \stackcell{18.40}{$11.3\times$} \\
Dream-7B & \stackcell{43.29}{$2.4\times$} & \stackcell{42.68}{$2.9\times$} & \stackcell{40.85}{$3.6\times$} & \stackcell{39.63}{$4.1\times$} & \stackcell{26.83}{$5.1\times$} & \stackcell{20.12}{$6.6\times$} & \stackcell{55.20}{$5.6\times$} & \stackcell{55.20}{$6.8\times$} & \stackcell{54.40}{$8.4\times$} & \stackcell{51.00}{$10.2\times$} & \stackcell{44.40}{$12.2\times$} & \stackcell{31.20}{$16.0\times$} & \stackcell{34.80}{$5.1\times$} & \stackcell{34.40}{$6.0\times$} & \stackcell{34.00}{$7.2\times$} & \stackcell{33.20}{$8.5\times$} & \stackcell{30.40}{$9.9\times$} & \stackcell{25.80}{$12.0\times$} \\
LLaDA-8B & \stackcell{33.54}{$6.2\times$} & \stackcell{33.54}{$7.9\times$} & \stackcell{33.54}{$10.4\times$} & \stackcell{31.10}{$13.1\times$} & \stackcell{29.88}{$15.0\times$} & \stackcell{25.61}{$18.8\times$} & \stackcell{40.60}{$4.8\times$} & \stackcell{40.60}{$6.2\times$} & \stackcell{41.00}{$8.3\times$} & \stackcell{40.80}{$10.3\times$} & \stackcell{36.80}{$12.9\times$} & \stackcell{27.80}{$16.8\times$} & \stackcell{27.80}{$4.8\times$} & \stackcell{28.40}{$5.8\times$} & \stackcell{28.40}{$7.0\times$} & \stackcell{27.20}{$8.3\times$} & \stackcell{24.80}{$9.9\times$} & \stackcell{21.60}{$12.0\times$} \\
\bottomrule
\end{tabular}
\normalsize
\caption{\textbf{Parallel decoding under confidence thresholding, $6.5$--$7$B trunks.}}
\label{fig:app-decode-7b-tau}
\end{figure}

\begin{figure}[!tp]
\centering
\fitwidth{\includegraphics{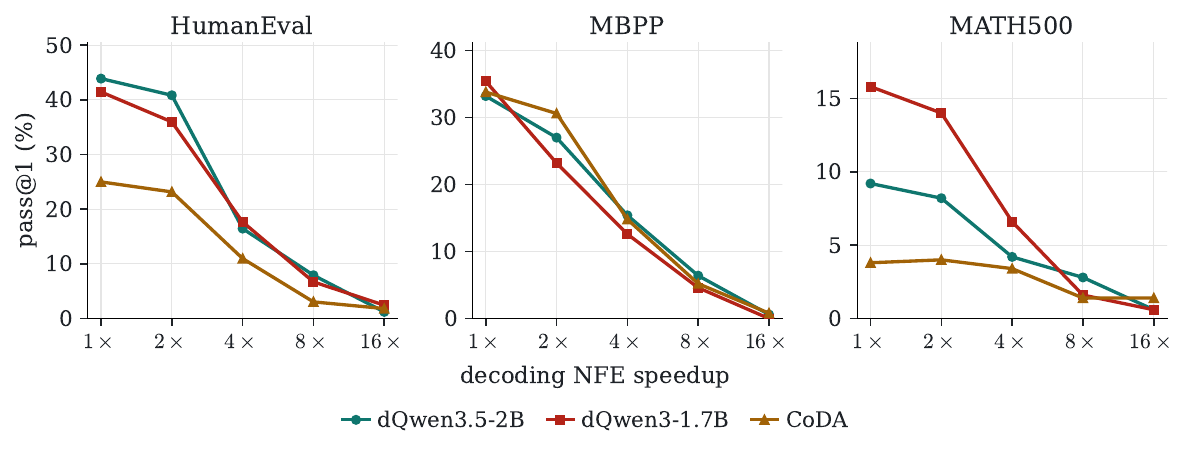}}
\\[6pt] \scriptsize
\setlength{\tabcolsep}{3pt}
\begin{tabular}{lrrrrrrrrrrrrrrr}
\toprule
 & \multicolumn{5}{c}{HumanEval} & \multicolumn{5}{c}{MBPP} & \multicolumn{5}{c}{MATH500} \\
\cmidrule(lr){2-6}\cmidrule(lr){7-11}\cmidrule(lr){12-16}
model & $1\times$ & $2\times$ & $4\times$ & $8\times$ & $16\times$ & $1\times$ & $2\times$ & $4\times$ & $8\times$ & $16\times$ & $1\times$ & $2\times$ & $4\times$ & $8\times$ & $16\times$ \\
\midrule
dQwen3.5-2B & 43.90 & 40.85 & 16.46 & 7.93 & 1.22 & 33.20 & 27.00 & 15.40 & 6.40 & 0.60 & 9.20 & 8.20 & 4.20 & 2.80 & 0.60 \\
dQwen3-1.7B & 41.46 & 35.98 & 17.68 & 6.71 & 2.44 & 35.40 & 23.20 & 12.60 & 4.60 & 0.00 & 15.80 & 14.00 & 6.60 & 1.60 & 0.60 \\
CoDA & 25.00 & 23.17 & 10.98 & 3.05 & 1.83 & 33.80 & 30.60 & 14.80 & 5.20 & 0.80 & 3.80 & 4.00 & 3.40 & 1.40 & 1.40 \\
\bottomrule
\end{tabular}
\normalsize
\caption{\textbf{Parallel decoding at fixed step budgets, ${\approx}1.4$B trunks.}}
\label{fig:app-decode-14b-fixed}
\end{figure}

\begin{figure}[!tp]
\centering
\fitwidth{\includegraphics{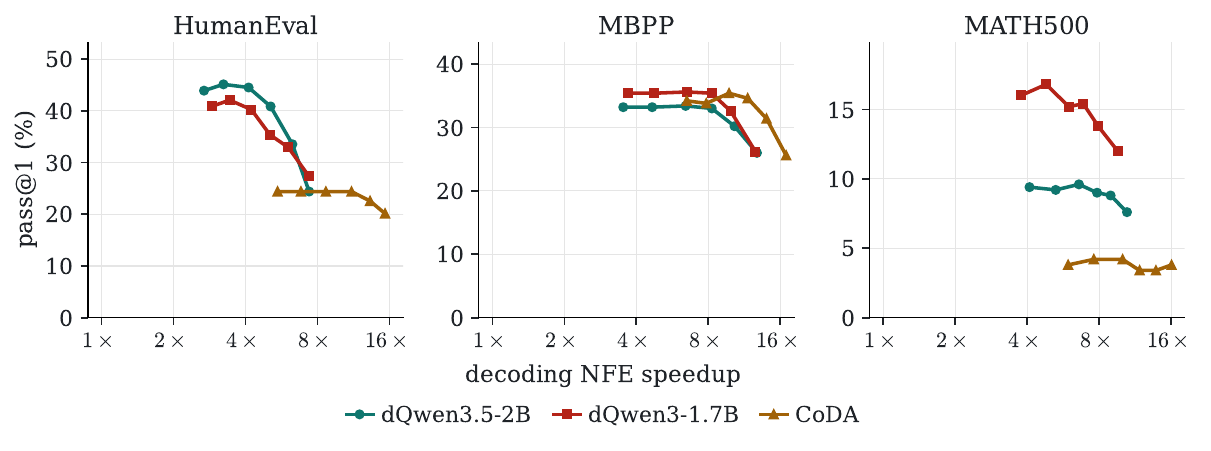}}
\\[6pt] \scriptsize
\setlength{\tabcolsep}{2pt}
\begin{tabular}{lrrrrrrrrrrrrrrrrrr}
\toprule
 & \multicolumn{6}{c}{HumanEval} & \multicolumn{6}{c}{MBPP} & \multicolumn{6}{c}{MATH500} \\
\cmidrule(lr){2-7}\cmidrule(lr){8-13}\cmidrule(lr){14-19}
model & 0.95 & 0.9 & 0.8 & 0.7 & 0.6 & 0.5 & 0.95 & 0.9 & 0.8 & 0.7 & 0.6 & 0.5 & 0.95 & 0.9 & 0.8 & 0.7 & 0.6 & 0.5 \\
\midrule
dQwen3.5-2B & \stackcell{43.90}{$2.7\times$} & \stackcell{45.12}{$3.2\times$} & \stackcell{44.51}{$4.1\times$} & \stackcell{40.85}{$5.1\times$} & \stackcell{33.54}{$6.3\times$} & \stackcell{24.39}{$7.4\times$} & \stackcell{33.20}{$3.5\times$} & \stackcell{33.20}{$4.7\times$} & \stackcell{33.40}{$6.4\times$} & \stackcell{33.00}{$8.3\times$} & \stackcell{30.20}{$10.3\times$} & \stackcell{26.00}{$12.8\times$} & \stackcell{9.40}{$4.1\times$} & \stackcell{9.20}{$5.3\times$} & \stackcell{9.60}{$6.6\times$} & \stackcell{9.00}{$7.8\times$} & \stackcell{8.80}{$8.9\times$} & \stackcell{7.60}{$10.5\times$} \\
dQwen3-1.7B & \stackcell{40.85}{$2.9\times$} & \stackcell{42.07}{$3.4\times$} & \stackcell{40.24}{$4.2\times$} & \stackcell{35.37}{$5.1\times$} & \stackcell{32.93}{$6.0\times$} & \stackcell{27.44}{$7.3\times$} & \stackcell{35.40}{$3.7\times$} & \stackcell{35.40}{$4.8\times$} & \stackcell{35.60}{$6.5\times$} & \stackcell{35.40}{$8.3\times$} & \stackcell{32.60}{$9.9\times$} & \stackcell{26.20}{$12.6\times$} & \stackcell{16.00}{$3.8\times$} & \stackcell{16.80}{$4.8\times$} & \stackcell{15.20}{$6.0\times$} & \stackcell{15.40}{$6.9\times$} & \stackcell{13.80}{$7.9\times$} & \stackcell{12.00}{$9.6\times$} \\
CoDA & \stackcell{24.39}{$5.4\times$} & \stackcell{24.39}{$6.8\times$} & \stackcell{24.39}{$8.7\times$} & \stackcell{24.39}{$11.1\times$} & \stackcell{22.56}{$13.3\times$} & \stackcell{20.12}{$15.3\times$} & \stackcell{34.20}{$6.5\times$} & \stackcell{33.80}{$7.8\times$} & \stackcell{35.40}{$9.8\times$} & \stackcell{34.60}{$11.7\times$} & \stackcell{31.40}{$14.0\times$} & \stackcell{25.60}{$16.9\times$} & \stackcell{3.80}{$5.9\times$} & \stackcell{4.20}{$7.6\times$} & \stackcell{4.20}{$10.0\times$} & \stackcell{3.40}{$11.8\times$} & \stackcell{3.40}{$13.8\times$} & \stackcell{3.80}{$16.1\times$} \\
\bottomrule
\end{tabular}
\normalsize
\caption{\textbf{Parallel decoding under confidence thresholding, ${\approx}1.4$B trunks.}}
\label{fig:app-decode-14b-tau}
\end{figure}

\begin{figure}[!tp]
\centering
\fitwidth{\includegraphics{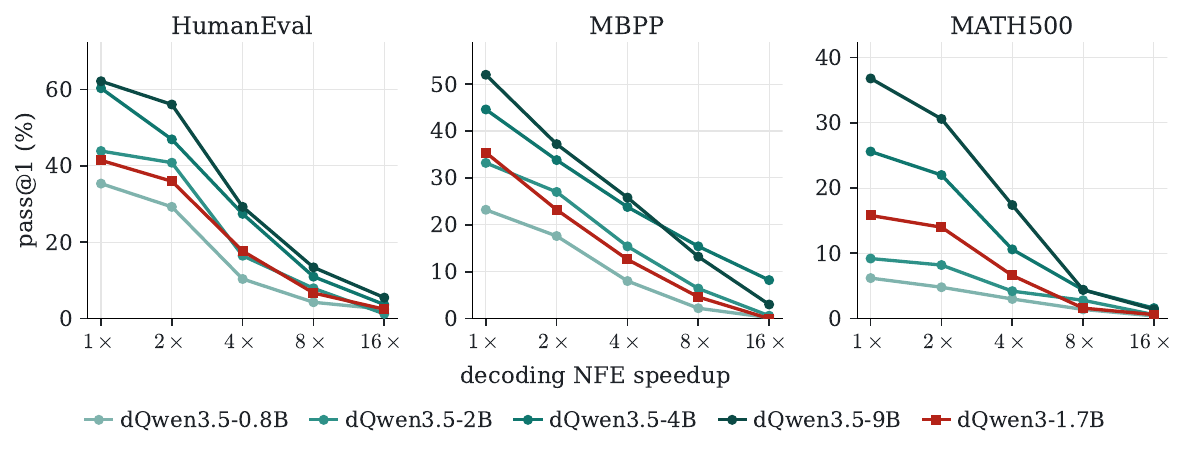}}
\\[6pt] \scriptsize
\setlength{\tabcolsep}{3pt}
\begin{tabular}{lrrrrrrrrrrrrrrr}
\toprule
 & \multicolumn{5}{c}{HumanEval} & \multicolumn{5}{c}{MBPP} & \multicolumn{5}{c}{MATH500} \\
\cmidrule(lr){2-6}\cmidrule(lr){7-11}\cmidrule(lr){12-16}
model & $1\times$ & $2\times$ & $4\times$ & $8\times$ & $16\times$ & $1\times$ & $2\times$ & $4\times$ & $8\times$ & $16\times$ & $1\times$ & $2\times$ & $4\times$ & $8\times$ & $16\times$ \\
\midrule
dQwen3.5-0.8B & 35.37 & 29.27 & 10.37 & 4.27 & 2.44 & 23.20 & 17.60 & 8.00 & 2.20 & 0.20 & 6.20 & 4.80 & 3.00 & 1.40 & 0.40 \\
dQwen3.5-2B & 43.90 & 40.85 & 16.46 & 7.93 & 1.22 & 33.20 & 27.00 & 15.40 & 6.40 & 0.60 & 9.20 & 8.20 & 4.20 & 2.80 & 0.60 \\
dQwen3.5-4B & 60.37 & 46.95 & 27.44 & 10.98 & 3.66 & 44.60 & 33.80 & 23.80 & 15.40 & 8.20 & 25.60 & 22.00 & 10.60 & 4.40 & 1.60 \\
dQwen3.5-9B & 62.20 & 56.10 & 29.27 & 13.41 & 5.49 & 52.00 & 37.20 & 25.80 & 13.20 & 3.00 & 36.80 & 30.60 & 17.40 & 4.40 & 1.40 \\
dQwen3-1.7B & 41.46 & 35.98 & 17.68 & 6.71 & 2.44 & 35.40 & 23.20 & 12.60 & 4.60 & 0.00 & 15.80 & 14.00 & 6.60 & 1.60 & 0.60 \\
\bottomrule
\end{tabular}
\normalsize
\caption{\textbf{Parallel decoding at fixed step budgets, our adapted models.}}
\label{fig:app-decode-family-fixed}
\end{figure}

\begin{figure}[!tp]
\centering
\fitwidth{\includegraphics{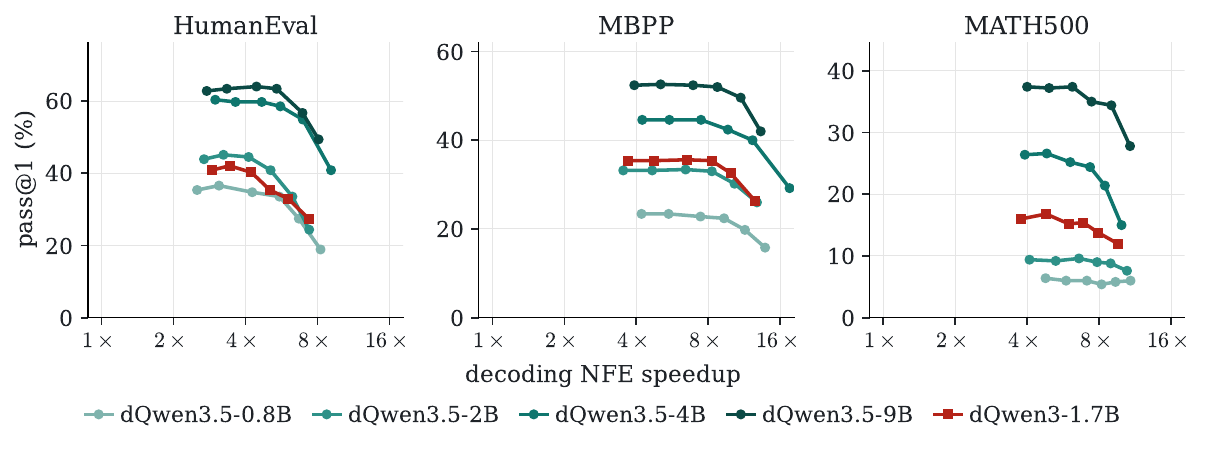}}
\\[6pt] \scriptsize
\setlength{\tabcolsep}{1.6pt}
\begin{tabular}{lrrrrrrrrrrrrrrrrrr}
\toprule
 & \multicolumn{6}{c}{HumanEval} & \multicolumn{6}{c}{MBPP} & \multicolumn{6}{c}{MATH500} \\
\cmidrule(lr){2-7}\cmidrule(lr){8-13}\cmidrule(lr){14-19}
model & 0.95 & 0.9 & 0.8 & 0.7 & 0.6 & 0.5 & 0.95 & 0.9 & 0.8 & 0.7 & 0.6 & 0.5 & 0.95 & 0.9 & 0.8 & 0.7 & 0.6 & 0.5 \\
\midrule
dQwen3.5-0.8B & \stackcell{35.37}{$2.5\times$} & \stackcell{36.59}{$3.1\times$} & \stackcell{34.76}{$4.3\times$} & \stackcell{33.54}{$5.5\times$} & \stackcell{27.44}{$6.7\times$} & \stackcell{18.90}{$8.2\times$} & \stackcell{23.40}{$4.2\times$} & \stackcell{23.40}{$5.5\times$} & \stackcell{22.80}{$7.4\times$} & \stackcell{22.40}{$9.3\times$} & \stackcell{19.80}{$11.4\times$} & \stackcell{15.80}{$13.8\times$} & \stackcell{6.40}{$4.8\times$} & \stackcell{6.00}{$5.8\times$} & \stackcell{6.00}{$7.1\times$} & \stackcell{5.40}{$8.2\times$} & \stackcell{5.80}{$9.4\times$} & \stackcell{6.00}{$10.8\times$} \\
dQwen3.5-2B & \stackcell{43.90}{$2.7\times$} & \stackcell{45.12}{$3.2\times$} & \stackcell{44.51}{$4.1\times$} & \stackcell{40.85}{$5.1\times$} & \stackcell{33.54}{$6.3\times$} & \stackcell{24.39}{$7.4\times$} & \stackcell{33.20}{$3.5\times$} & \stackcell{33.20}{$4.7\times$} & \stackcell{33.40}{$6.4\times$} & \stackcell{33.00}{$8.3\times$} & \stackcell{30.20}{$10.3\times$} & \stackcell{26.00}{$12.8\times$} & \stackcell{9.40}{$4.1\times$} & \stackcell{9.20}{$5.3\times$} & \stackcell{9.60}{$6.6\times$} & \stackcell{9.00}{$7.8\times$} & \stackcell{8.80}{$8.9\times$} & \stackcell{7.60}{$10.5\times$} \\
dQwen3.5-4B & \stackcell{60.37}{$3.0\times$} & \stackcell{59.76}{$3.6\times$} & \stackcell{59.76}{$4.7\times$} & \stackcell{58.54}{$5.6\times$} & \stackcell{54.88}{$6.9\times$} & \stackcell{40.85}{$9.1\times$} & \stackcell{44.60}{$4.2\times$} & \stackcell{44.60}{$5.5\times$} & \stackcell{44.60}{$7.5\times$} & \stackcell{42.40}{$9.7\times$} & \stackcell{40.00}{$12.3\times$} & \stackcell{29.20}{$17.5\times$} & \stackcell{26.40}{$3.9\times$} & \stackcell{26.60}{$4.8\times$} & \stackcell{25.20}{$6.1\times$} & \stackcell{24.40}{$7.3\times$} & \stackcell{21.40}{$8.5\times$} & \stackcell{15.00}{$9.9\times$} \\
dQwen3.5-9B & \stackcell{62.80}{$2.8\times$} & \stackcell{63.41}{$3.3\times$} & \stackcell{64.02}{$4.4\times$} & \stackcell{63.41}{$5.4\times$} & \stackcell{56.71}{$6.9\times$} & \stackcell{49.39}{$8.1\times$} & \stackcell{52.40}{$3.9\times$} & \stackcell{52.60}{$5.1\times$} & \stackcell{52.40}{$6.9\times$} & \stackcell{52.00}{$8.7\times$} & \stackcell{49.60}{$10.9\times$} & \stackcell{42.00}{$13.3\times$} & \stackcell{37.40}{$4.0\times$} & \stackcell{37.20}{$4.9\times$} & \stackcell{37.40}{$6.2\times$} & \stackcell{35.00}{$7.4\times$} & \stackcell{34.40}{$9.0\times$} & \stackcell{27.80}{$10.8\times$} \\
dQwen3-1.7B & \stackcell{40.85}{$2.9\times$} & \stackcell{42.07}{$3.4\times$} & \stackcell{40.24}{$4.2\times$} & \stackcell{35.37}{$5.1\times$} & \stackcell{32.93}{$6.0\times$} & \stackcell{27.44}{$7.3\times$} & \stackcell{35.40}{$3.7\times$} & \stackcell{35.40}{$4.8\times$} & \stackcell{35.60}{$6.5\times$} & \stackcell{35.40}{$8.3\times$} & \stackcell{32.60}{$9.9\times$} & \stackcell{26.20}{$12.6\times$} & \stackcell{16.00}{$3.8\times$} & \stackcell{16.80}{$4.8\times$} & \stackcell{15.20}{$6.0\times$} & \stackcell{15.40}{$6.9\times$} & \stackcell{13.80}{$7.9\times$} & \stackcell{12.00}{$9.6\times$} \\
\bottomrule
\end{tabular}
\normalsize
\caption{\textbf{Parallel decoding under confidence thresholding, our adapted models.}}
\label{fig:app-decode-family-tau}
\end{figure}

\Cref{fig:app-decode-7b-fixed,fig:app-decode-7b-tau} tabulate the two decoding schemes from~\cref{sec:eval-decode}, while \cref{fig:app-decode-14b-fixed,fig:app-decode-14b-tau} extend the comparison to the ${\approx}1.4$B trunks and \cref{fig:app-decode-family-fixed,fig:app-decode-family-tau} cover the full \dqfam{} family.
We evaluate fixed budgets from 64 to 1024 NFEs and confidence thresholds from $\tau=0.5$ to $\tau=0.95$.
Across scales, the hybrid models retain meaningful accuracy as the NFE budget falls, with larger models generally supporting more aggressive parallel decoding.

\FloatBarrier

\clearpage
\subsection{Forward-pass throughput}
\label{app:throughput}

\begin{figure}[!ht]
\centering
\fitwidth{\includegraphics{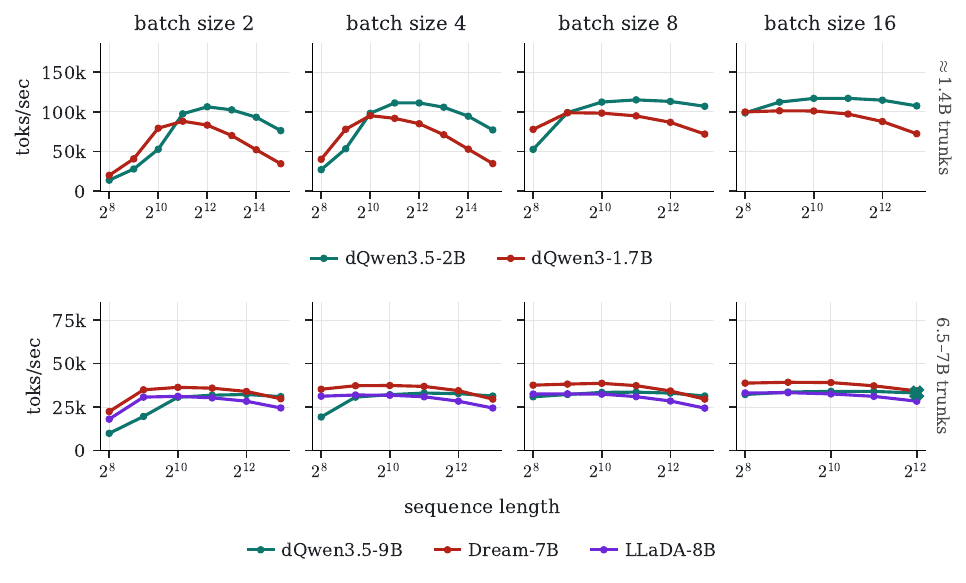}}
\caption{\textbf{Forward-pass throughput.}
Whole-model throughput on one NVIDIA GH200 using the standard fast-path kernels.
At the $\approx$1.4B trunk scale, the hybrid model outperforms its full-attention control at longer sequences despite having more total parameters; at 6.5--7B, it remains competitive with other full-attention DLMs.
}
\label{fig:throughput}
\end{figure}

Because $3/4$ of Qwen3.5's sequence-processing layers are recurrent GDNs, we also ask whether they slow each model forward pass.
We measure whole-model throughput across sequence lengths and batch sizes on one NVIDIA GH200 using the standard fast-path kernels: PyTorch flash SDPA for attention and the FLA GDN fast path with \texttt{causal\_conv1d} for hybrid models (\cref{fig:throughput}).
We omit CoDA and Dream-Coder-7B because they share backbone architectures with \dqctrl{}-1.7B and Dream-7B, respectively.

At the $\approx$1.4B trunk scale, \dqfam{}-2B is slower at short sequences but overtakes the full-attention control as sequence length grows.
At the 6.5--7B scale, \dqfam{}-9B shows the same trend: it begins slower, but approaches the full-attention models at longer sequences.
Thus, causal recurrence does not require slow forward passes; with fast-path kernels, hybrid models remain competitive as sequence length grows while using full attention in only a minority of its layers.

\FloatBarrier



\end{document}